\documentclass[10pt,journal,compsoc]{IEEEtran}
\usepackage{times}
\usepackage{epsfig}
\usepackage{graphicx}
\usepackage{amsmath}
\usepackage{amssymb}
\usepackage{hyperref}       % hyperlinks
\usepackage{url,subfig}            % simple URL typesetting
\usepackage{booktabs}       % professional-quality tables
\usepackage{amsfonts}       % blackboard math symbols
\usepackage{nicefrac}       % compact symbols for 1/2, etc.
\usepackage{microtype}      % microtypography
\usepackage{xcolor}         % colors
\usepackage{bm,float}
\usepackage{makecell}
\usepackage{graphicx}
\usepackage{multirow}
\usepackage{mathrsfs}
\usepackage{amsthm}
\usepackage{algorithm} 
\usepackage{algorithmic}
\usepackage{lineno}
\usepackage{todonotes}
\usepackage[english]{babel}

\newtheorem{theorem}{Theorem}

\newtheorem{remark}{Remark}

\ifCLASSOPTIONcompsoc
  \usepackage[nocompress]{cite}
\else
  \usepackage{cite}
\fi

\ifCLASSINFOpdf
\else
\fi
\begin{document}
\title{COTReg: Coupled Optimal Transport based Point Cloud Registration}

\author{Guofeng Mei, Xiaoshui Huang, Litao Yu, Jian Zhang, Qiang Wu, \IEEEmembership{Senior Member,~IEEE,} and Mohammed Bennamoun, \IEEEmembership{Senior Member,~IEEE}, 
% 	<-this stops a space
\IEEEcompsocitemizethanks{\IEEEcompsocthanksitem Guofeng Mei, Litao Yu, Qiang Wu, and Jian Zhang are with the Faculty of Engineering and Information Technology, University of Technology Sydney, Sydney NSW 2007, Australia. (e-mail: {litao.yu; qiang.wu; jian.zhang}@uts.edu.au; guofeng.mei@student.uts.edu.au).

\IEEEcompsocthanksitem Xiaoshui Huang is with the Image X Institute of Faculty of Medicine and Health, University of Sydney, Sydney NSW 2015, Australia.
% \IEEEcompsocthanksitem Hao Tang is with the Department of Information Technology and Electrical Engineering, ETH Zurich,  Zurich 8092, Switzerland.
\IEEEcompsocthanksitem Mohammed Bennamoun is with the Department of Computer Science and Software Engineering, the University of Western Australia, WA 6009, Australia.
% \IEEEcompsocthanksitem Nicu Sebe is with the Department of Information Engineering and Computer Science (DISI), University of Trento, Italy.
}% <-this % stops a space
\thanks{Corresponding author: Jian Zhang (Jian.Zhang@uts.edu.au).}
}
\IEEEtitleabstractindextext{%
\begin{abstract}
Generating a set of high-quality correspondences or matches is one of the most critical steps in point cloud registration. This paper proposes a learning framework COTReg by simultaneously considering point-wise and structural matchings to estimate correspondences in a coarse-to-fine manner. Specifically, we transform the two matchings into a Wasserstein distance-based and a Gromov-Wasserstein distance-based optimizations, respectively. The task of establishing the correspondences thus can be naturally reshaped to a coupled optimal transport problem. Furthermore, we design an overlap attention module consisting primarily of transformer layers and train it to predict the probability (overlap score) that each point lies in the overlapping region. The overlap score then guides the procedure of correspondences prediction. We conducted comprehensive experiments on 3DMatch, KITTI, and 3DCSR benchmarks showing the state-of-art performance of the proposed method.
\end{abstract}

}

\maketitle

% is not selected <OR> if conference mode is selected - because compsoc
% conference papers position the abstract like regular (non-compsoc)
\IEEEdisplaynontitleabstractindextext

% \fi
\IEEEpeerreviewmaketitle

\ifCLASSOPTIONcompsoc
\IEEEraisesectionheading{\section{Introduction}\label{sec:introduction}}
\else
\section{Introduction}
\label{sec:introduction}
\fi

\IEEEPARstart{P}{oint} cloud registration is to align two or more 3D point clouds acquired from various views, platforms, or at different times into a unified coordinate system \cite{wang2019prnet}. This technique is the cornerstone of many 3D computer vision applications, such as 3D reconstruction \cite{wang2017feature}, augmenting reality \cite{borrmann2018large}, autonomous driving \cite{chen20203d,nagy2018real,wang2019robust}, cancer radiotherapy \cite{ma2018point,li2019noninvasive} and robotics \cite{saputra2018casualty,pomerleau2015review}. 

Recently, various types of deep learning-based point cloud registration approaches have achieved outstanding performance. Some of these methods show that combining conventional optimization approaches and current deep neural networks obtains better accuracies than traditional optimization methods \cite{choy2020deep}.  Algorithms based on correspondences occupy a significant proportion of deep learning-based point cloud registration methods. State-of-the-art correspondence-based pipelines commonly consist of the following stages \cite{bai2021pointdsc}: \textit{feature extraction, correspondence prediction, outlier rejection}, and \textit{transformation estimation}. Such correspondence-based approaches mainly focus on improving registration performance by extracting highly discriminative features \cite{yew20183dfeat,gojcic2019perfect,huang2021predator,fu2021robust} or removing outlier correspondences \cite{choy2020deep,bai2021pointdsc,pais20203dregnet}.  The correspondence prediction is also a critical stage since the estimation of the rigid motion parameters depends on the correct correspondences. However,  very few works have been devoted to improving correspondence prediction algorithms \cite{choy2020deep}.   Therefore, this paper focuses on developing a correspondence prediction algorithm to obtain more accurate correspondences for pairwise point cloud registration.

\begin{figure}[t]
	\includegraphics[width=1\linewidth]{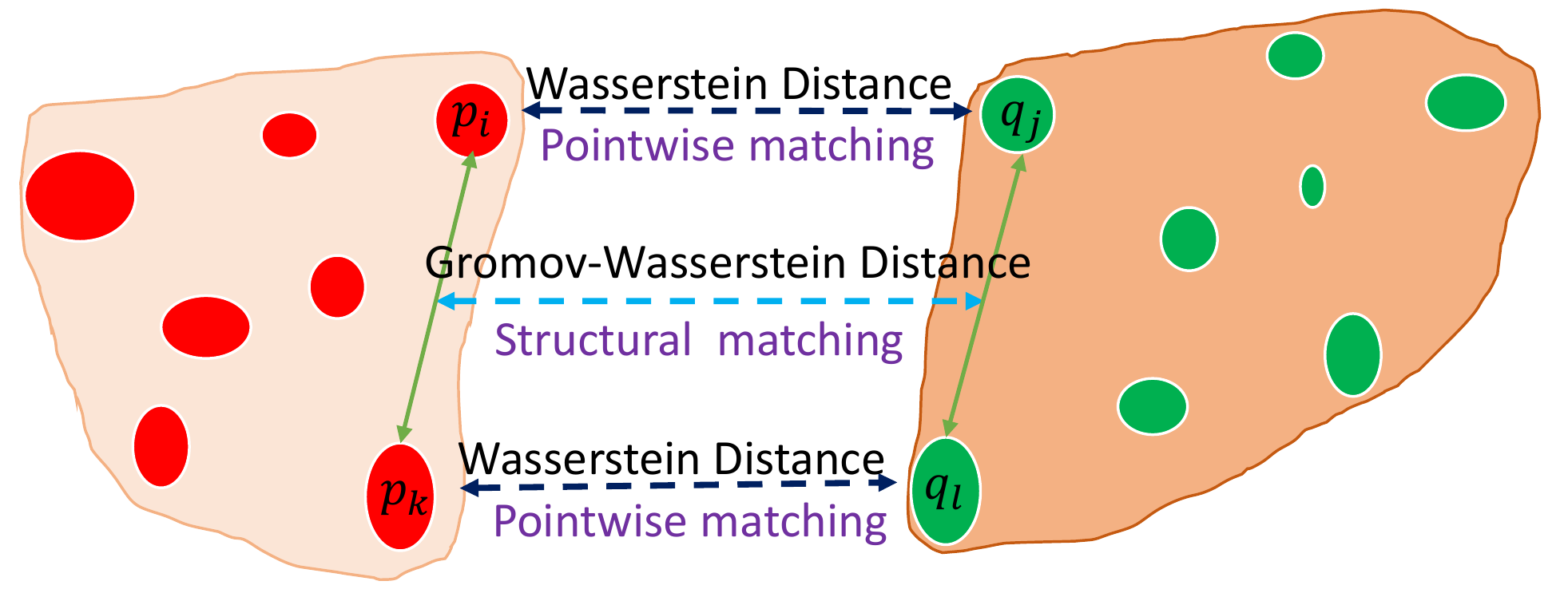}
	\caption{A paradigm shows the key concept of the proposed correspondence prediction algorithm. It consists of the pointwise matching based on the feature similarities, such as $\bm{p}_i$ and $\bm{q}_j$, and structural matching based on the geometric similarities between two pairs of points, such as $\{\bm{p}_i$, $\bm{p}_k\}$ and $\{\bm{q}_j,\bm{q}_l\}$. The size of each point represents its overlap score.}
	\label{eq:fram1}
\end{figure}

In learning-based 3D point registration, pointwise matching is often used to establish matches between two point clouds \cite{zeng20173dmatch}. The core idea of pointwise matching is that a pair of points between the source and target point clouds having the most similar feature representations are identified as the corresponding points. However, the putative correspondences produced by the pointwise matching contain many false matches and outliers \cite{li2020robust}. For example, the feature matching recall of recent FCGF \cite{choy2019fully} drops to 80\% when the inlier ratio is set to 0.2, meaning at least 20\% correspondences contain more than 80\% outliers \cite{choy2019fully}. We observe that the following two factors will cause false matches and outliers: first of all, some inliers are assigned to outliers because we treat the inliers and outliers equally in the correspondence prediction stage \cite{huang2021predator}; second, only considering pointwise matching is insufficient to find the correct correspondences due to the ambiguous and repeated patterns in the 3D point clouds \cite{fu2021robust}. To solve the above problems, we introduce the overlap scores, which act as auxiliary information to assist in correspondence prediction. When combining overlap scores with pointwise matching, it induces an optimal transport problem \cite{peyre2019computational}, i.e., Wasserstein distance (\textbf{WD}) \cite{peyre2019computational}. Furthermore, inspired by the success of structural information (edge length) in graph matching \cite{fey2020deep,zanfir2018deep}, we also consider it as an additional constraint to produce correspondences for registration. In our case, the structure is defined as the distance between two points within a point cloud in both the Euclidean and feature spaces, so the structural matching is formulated by comparing the geometric difference based on the Gromov-Wasserstein distance (\textbf{GWD}) \cite{peyre2019computational}. The pointwise and structural matchings are illustrated in Figure \ref{eq:fram1}. By integrating the pointwise and structural matchings into a unified learning framework, we formulate the problem of establishing the correspondences of point cloud registration as a coupled optimal transport problem. It can be easily integrated into deep neural networks for end-to-end training. Our method is based on the intuition that incorporating both pointwise and structural matchings into producing correspondences can resolve the ambiguity issue, achieving better performance than using either one of the two matchings. We also design a positional encoding scheme that assigns intrinsic geometric properties to per-point features to enhance distinctions among point features in indistinctive regions. Specifically, the coupled optimal transport can incorporate both the pointwise and the structural matchings with the overlap scores acting as empirical distributions to effectively improve the accuracy of the generated correspondences in 3D point cloud registration. To reduce the complexity of the matching process, our model adopts the coarse-to-fine mechanism, a hierarchical matching strategy to generate the correspondences.

To the best of our knowledge, we are the first to apply coupled optimal transport to predict correspondences for point cloud registration. The contribution of this paper is summarized as follows:

\begin{itemize}
	\item  We incorporate overlap scores, pointwise and structural matchings in a joint model based on coupled optimal transport to produce correspondences in a coarse-to-fine manner.
	\item We transform the pointwise matching into Wasserstein distance-based optimization to generate correspondences.  
	\item We cast the structural matching into Gromov-Wasserstein distance-based optimization to predict matches. The structural matching is formulated by considering the structural difference in both Euclidean and feature spaces. 
	\item We introduce an overlap attention module to learn point-wise features and overlap scores.
	\item Comprehensive experiments on indoor and outdoor, synthetic, and cross-source point clouds demonstrate that the proposed method can effectively improve the accuracy of 3D point cloud registration.
\end{itemize}

%The remainder of this paper is organized as follows. Section \ref{sec:rw} reviews related works. Section \ref{sec:probf} gives the problem formulation of point cloud registration. Section \ref{sec:exp} presents the evaluation methodology with descriptions on datasets, metrics, considered terms for evaluation. Section 5 presents the evaluation
%results and relevant explanations. Section 6 gives a summary
%and discussion for each method based on the evaluation
%The conclusions are finally drawn in Section \ref{seq:con}.

\section{Related Work}\label{sec:rw}
We review both traditional and deep learning-based point cloud registration methods. Since optimal transport is a major component in the 3D correspondence prediction of the proposed learning framework, we also review the relevant works.

\subsection{Traditional Point Cloud Registration}
The best-known traditional registration algorithm is Iterative Closest Point (ICP) \cite{besl1992method}, which alternates between rigid motion estimation and the correspondences searching \cite{wang2019deep,huang2017coarse} by solving a $L_2 $-optimization. However, ICP converges to spurious local minima due to the non-convexity. Based on this algorithm, many variants have been proposed. For example, the Levenberg-Marquardt ICP \cite{fitzgibbon2003robust} uses a Levenberg-Marquardt algorithm to produce a transformation by replacing the singular value decomposition with gradient descent and Gaussian-Newton approaches, accelerating convergence while ensuring high accuracy. Go-ICP \cite{yang2015go} solves the point cloud alignment problem globally by using a Branch-and-Bound (BnB) optimization framework without prior information on correspondence or transformation. RANSAC-like algorithms are widely used for robust finding of the correct correspondences for registration \cite{li2021point}. FGR \cite{zhou2016fast} optimizes a Geman-McClure cost-induced correspondence-based objective function in a graduated non-convex strategy and achieves high performance. TEASER \cite{yang2020teaser} reformulates the registration problem as an intractable optimization and provides readily checkable conditions to verify the optimal solution. However, most existing methods still face challenges when point clouds contain a mixture of noise, outlier, and partial overlap \cite{zhang2020deep}.  In contrast, the learning-based methods tend to have strong robustness \cite{ao2021spinnet}.

\subsection{Deep Learning-Based Point Cloud Registration}
Deep point cloud registration methods can be commonly classified into correspondences-free methods and correspondences-based methods \cite{zhang2020deep}. The core idea of correspondence-free registration approaches \cite{aoki2019pointnetlk,huang2020feature} is to estimate the transformation by minimizing the difference between global features extracted from two input point clouds. However, the global features rely on the whole points of a point cloud, leading to a major limitation that correspondences-free approaches are inapplicable to real scenes with partial overlaps \cite{choy2020deep,zhang2020deep}. 
On the contrary, correspondence-based methods extract the per-point or per-patch embeddings of the two input point clouds to generate a mapping and estimate a rigid transformation \cite{zhang2020deep}. For instance, DCP \cite{wang2019deep} employs DGCNN \cite{wang2019dynamic} for feature extraction and an attention module to generate soft matching pairs. FCGF \cite{choy2019fully} uses $1\times 1 \times 1$ kernel to extract compact metric features for geometric correspondence. RGM \cite{fu2021robust} considers information exchange between two point clouds to extract discriminative features for point-wise matching. Predator \cite{huang2021predator} and PRNet \cite{wang2019prnet} pay attention to the detection of points in the overlap region and use these features of the detected points to generate matches. DGR \cite{choy2020deep} proposes a 6-dimensional convolutional network architecture for inlier likelihood prediction and estimates the transformation via a weighted Procrustes module. 3DRegNet \cite{pais20203dregnet} uses an inlier prediction model to estimate the inliers. PointDSC \cite{bai2021pointdsc} integrates pairwise similarity as a constraint to improve correspondence accuracy. RPMNet \cite{yew2020rpm} and \cite{dang2020learning} propose methods to solve the point cloud partial visibility by integrating the Sinkhorn algorithm into a network to get soft correspondences from local features. Soft correspondences can increase the robustness of registration accuracy. IDAM \cite{li2019iterative} incorporates both geometric and distance features into the iterative matching process. Some works \cite{yew20183dfeat,gojcic2019perfect,el2021unsupervisedr,ding2019deepmapping} also focus on alleviating the dependence on the manual annotation in the process of extracting features. Most of these methods generate correspondences by applying the point feature similarities then directly match the highest response \cite{li2019iterative}.  This strategy has two obvious limitations. \textbf{First} of all, some inliers are assigned to outliers, since inliers and outliers are treated equally in the correspondence prediction stage \cite{huang2021predator}. \textbf{Second}, the one-shot matching may fail because there are multiple possible correspondences due to randomness \cite{li2019iterative}, while the per-point feature itself is not powerful enough to distinguish the geometric properties in point clouds \cite{fu2021robust}. Based on such observations, some additional constraints should be imposed to obtain high-quality correspondences. Inspired by edge similarity in graph matching \cite{fey2020deep,zanfir2018deep}, the matched pairs between two point clouds hold an equal length constraint in both Euclidean and feature spaces, we propose a coupled optimal transport-based framework that incorporates overlap scores into pointwise matching as well as structural matching to generate more accurate correspondences.  We also design a positional encoding scheme that assigns intrinsic geometric properties to per-point features to enhance distinctions among point features in indistinctive regions.

%In conclusion, only relying on pointwise matching is insufficient to find the correct correspondences due to ambiguous and repeated patterns in the 3D acquisition point clouds. Additional constraints should be considered to generate high-quality correspondences. 

\subsection{Optimal Transport}
Optimal transport (OT) \cite{peyre2019computational} is a method to exploit the best assignment between two general objects, which is widely applied in many machine learning applications \cite{chapel2020partial}. OT provides a way to predict the correspondences and measure the similarity between two distributions. The distance based on OT is called the Wasserstein distances (WD), which has been used in various computer vision tasks \cite{solomon2015convolutional}. For instance, Su et al. \cite{su2015optimal} employed the Wasserstein distance to deal with the 3D shape matching and surface registration problem. Dang et al. \cite{dang2020learning} applied Wasserstein distances to handle 3D point cloud registration. Gromov-Wasserstein (GWD) \cite{peyre2019computational} extends WD by computing couplings between metric measure spaces. Unlike calculating the distance between two entity sets as in WD, GWD can be utilized to calculate the structure distance within each domain. The GWD has been used for shape analysis \cite{solomon2016entropic}, graph matching \cite{xu2019gromov}, etc. The properties of WD and GWD that focus, respectively, on features and structure information motivate us to utilize WD to build pointwise matching and GWD to build structural matching. Contrasting with previous correspondence prediction algorithms for point-cloud registration, we treat point clouds as probability distributions, i.e., overlap scores are embedded in a specific metric space based on the similarities of features and structures.  The correspondence prediction is then used to compute a coupled optimal transport distance between two probability distributions that can endow pointwise and structural matchings with overlap scores to generate more accurate correspondences.

\section{Problem Formulation} \label{sec:probf}
Before introducing our method, we explain the formulation and notation of the problem. We consider two point clouds $\bm{\mathcal{P}}=\{\bm{p}_i\in \mathbb{R}^3|i=1,\cdots,N\}$ and $\bm{\mathcal{Q}}=\{\bm{q}_i\in \mathbb{R}^3|i=1,\cdots,M\}$, with their associated features $\bm{\mathcal{F}}_p=\{\bm{f}_{\bm{p}_i} \in \mathbb{R}^d|i=1,\cdots,N\}$ and $\bm{\mathcal{F}}_q=\{\bm{f}_{\bm{q}_i}\in \mathbb{R}^d|i=1,\cdots, M\}$, respectively. The registration problem is to recover a rigid transformation $\bm{T}$ with rotation $\bm{R}\in SO(3)$ and translation $\bm{t}\in \mathbb{R}^3$ that aligns the source set $\bm{\mathcal{P}}$ to the target set $\bm{\mathcal{Q}}$. Obviously $\bm{T}$ can only be determined from the data in overlapping areas of $\bm{\mathcal{P}}$ and $\bm{\mathcal{Q}}$, if both sets have sufficient overlaps. We define an assignment matrix $\Gamma\in\mathbb{R}^{N\times M}$ with elements $\Gamma_{ij}\in \{0, 1\}$ to represent the matching confidence between the point $\bm{p}_i$ and $\bm{q}_j$, where each element satisfies
\begin{equation}\label{eq:gamma}
	\Gamma_{ij} = 
	\begin{cases}
		1, & \text{if point} ~\bm{p}_i ~ \text{corresponds to} ~ \bm{q}_j\\
		0, & \text{otherwise}
	\end{cases}.
\end{equation}
Let us first consider the case where the overlapping region is given, and we use the overlap scores to depict the overlapping region. The overlap score of the point $\bm{p}_i$ is defined as 
\begin{equation}\label{eq:os0}
	\bm{\mu}_{\bm{p}_i} = 
	\begin{cases}
		1, & \text{if point} ~\bm{p}_i ~ \text{is an inlier of} ~ \bm{\mathcal{P}}\\
		0, & \text{otherwise}
	\end{cases},
\end{equation}
where $\bm{\mu}_{\bm{q}_j}$ is defined similarly. Let $\bm{\mu}_p = \{\bm{\mu}_{\bm{p}_i}|i=1,\cdots, N\}$ and $\bm{\mu}_q = \{\bm{\mu}_{\bm{q}_j}|j=1,\cdots, M\}$. Usually, it is unlikely to determine whether a point is in overlap regions. Thus we relax the overlap scores to $\bm{\mu}_{\bm{x}}\in[0, 1]$ and use a neural network to estimate overlap scores. The details of the overlap score prediction module are illustrated in Sec.~\ref{sec:method}. When considering the overlap scores, the registration task can be cast to solve the following problem:
\begin{equation}\label{eq:reg0}
	\begin{aligned}
		& \min_{\Gamma, \bm{R}, \bm{t}}\sum_{i=1}^N\sum_{j=1}^M\Gamma_{ij}\|\bm{R}\bm{p}_i + \bm{t}-\bm{q}_j\|_2, \\
		& \mbox{s.t.}~\sum_{j=1}^{M}\Gamma_{ij}=\bm{\mu}_{\bm{p}_i}, \sum_{i=1}^{N}\Gamma_{ij}=\bm{\mu}_{\bm{q}_j}, \Gamma_{ij}\in \{0, 1\}.
	\end{aligned}
\end{equation}
The three constraints enforce $\Gamma$ to be a permutation matrix. If we know the optimal rigid transformation $\{\bm{R}, \bm{t}\}$, then the assignment matrix $\Gamma$ can be recovered from Eq. \eqref{eq:reg0}. By contrary, given the assignment matrix $\Gamma$, we generate correspondences $\mathcal{M} = \{(\bm{p}_i, \bm{q}_j)|j = \arg\max_{k}\Gamma_{ik}\}$. $\mathcal{M}$ can be directly leveraged by RANSAC \cite{fischler1981random} or SVD to estimate the transformation. 

The following notation will be used throughout the paper. $\bm{p}_i\rightarrow\bm{q}_j$ indicates a correspondence of $\bm{p}_i$ and $\bm{q}_j$. $\{\bm{p}_i, \bm{p}_k\}$ represents a pair. $\left<\cdot,\cdot\right>$ is an inner product operator. The Kullback-Leibler ($\mathcal{KL}$) divergence between two non-negative vectors $\bm{a}=\left(a_1,a_2,\cdots, a_n\right)$ and $\bm{b}=\left(b_1,b_2,\cdots, b_n\right)$ is defined as
\begin{equation}\label{eq:kl}
	\mathcal{KL}\left(\bm{a}|\bm{b}\right)=\sum_{i=1}^n \left(a_i\log\left(\frac{a_i}{b_i}\right)-a_i + b_i\right),	 
\end{equation}
with the convention $ 0 \log 0 = 0$.

\section{Proposed optimal transport-based point cloud registration}\label{sec:method}
\begin{figure}[htbp]
	\centering
	\includegraphics[width=1.0\linewidth]{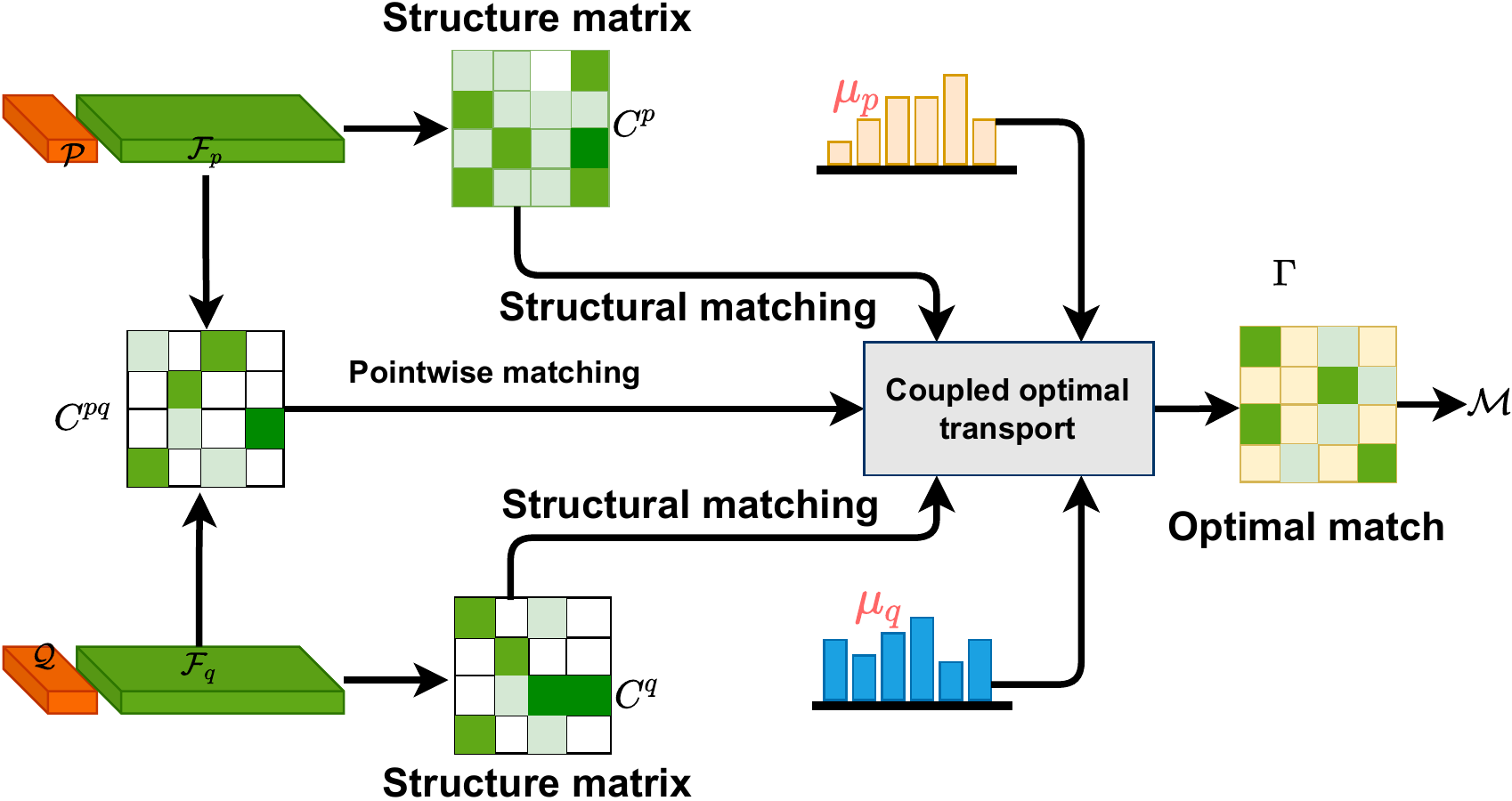}
	\caption{\textbf{Overview of the correspondence prediction.} Point clouds $\bm{\mathcal{P}}$ and $\bm{\mathcal{Q}}$, with their features $\bm{\mathcal{F}}_p$ and $\bm{\mathcal{F}}_q$, and the overlap scores $\bm{\mu}_p, \bm{\mu}_q$.  $\bm{C}^{pq}$ is the cross distance matrix. $\bm{C}^p$ and $\bm{C}^q$ represent the structure matrices. The assignment matrix $\Gamma$ is predicted by solving a coupled optimal transport problem. $\bm{\mathcal{P}}$ and $\bm{\mathcal{Q}}$ have $N$ and $M$ points, respectively. $\mathcal{M}$ represents a set of estimated correspondences.} \label{fig:fram}
\end{figure}
In this paper, we propose a new method to generate correspondences by jointly considering pointwise and structural matchings, as illustrated in Figure \ref{fig:fram}. We first use $\bm{\mathcal{F}}_p$ and the coordinates of $\bm{\mathcal{P}}$ to calculate $\bm{C}^p$. Similarly, $\bm{\mathcal{F}}_q$ and the coordinates of $\bm{\mathcal{Q}}$ are used to calculate the $\bm{C}^q$. After that, $\bm{C}^p$, $\bm{C}^q$, the overlap scores $\bm{\mu}_p$ and $\bm{\mu}_q$ are used for structural matching based on the Gromov-Wasserstein distance. $\bm{\mathcal{F}}_p$ and $\bm{\mathcal{F}}_q$ are used to calculate $\bm{C}^{pq}$. Then $\bm{C}^{pq}$ and overlap scores $\bm{\mu}_p$ and $\bm{\mu}_q$ are used for pointwise matching based on the Wasserstein distance. Finally, we couple Wasserstein distance and Gromov-Wasserstein distance in a mutually-beneficial way by sharing the assignment matrix $\Gamma$, which is estimated via the coupled optimal transport algorithm.  $\bm{\mathcal{P}}$ and $\bm{\mathcal{Q}}$ have $N$ and $M$ points, respectively. The correspondences $\mathcal{M}$ are finally obtained based on $\mathcal{M} = \{(\bm{p}_i, \bm{q}_j)|j = \arg\max_{k}\Gamma_{ik}\}$.
Next, we introduce how to use the coupled optimal transport to predict the correspondences.

\subsection{Coupled Optimal Transport-Based Correspondence Prediction} \label{subs:coot}
As is mentioned above, only considering the pointwise matching is not sufficient to find the accurate correspondences due to the ambiguous and repeated patterns in the 3D acquisition point clouds. Therefore, we exploit both pointwise and structural matchings to indicate the correspondence between source and target point clouds.

For point-wise matching, the cross-distance matrix $\bm{C}^{pq}\in\mathbb{R}^{N\times M}$ is derived based on the feature space between point cloud $\bm{\mathcal{P}}$ and $\bm{\mathcal{Q}}$ with each element satisfying
\begin{equation}
	\bm{C}^{pq}_{ij}=\mathcal{D}_f\left(\bm{f}_{\bm{p}_i}, \bm{f}_{\bm{q}_j}\right), 0\leq i \leq N, 0\leq j \leq M,
\end{equation}
where $\mathcal{D}_f\left(\bm{f}_{\bm{p}_i},\bm{f}_{\bm{q}_j}\right)=\|\frac{\bm{f}_{\bm{p}_i}}{\|\bm{f}_{\bm{p}_i}\|_2}-\frac{\bm{f}_{\bm{q}_j}}{\|\bm{f}_{\bm{q}_j}\|_2}\|_2$ represents the distance between $\bm{f}_{\bm{p}_i}$ and $\bm{f}_{\bm{q}_j}$.

To construct the structural matching, we first calculate the discrepancy (structure) matrix $\bm{C}^p\in\mathbb{R}^{N\times N}$ of the point pairs within $\bm{\mathcal{P}}$ in both Euclidean and feature spaces with the elements related to $\{\bm{p}_i, \bm{p}_k\}$ satisfying  
\begin{equation}\label{eq:mp}
	\bm{C}^p_{ik}=\lambda\underbrace{\mathcal{D}_e\left(\bm{p}_i,\bm{p}_k\right)}_{Euclidean}+(1-\lambda)\underbrace{\mathcal{D}_f\left(\bm{f}_{\bm{p}_i},\bm{f}_{\bm{p}_k}\right)}_{Feature},
\end{equation}
where $\mathcal{D}_e\left(\cdot, \cdot\right)$ is a function related to distances between two points in Euclidean space satisfying $\mathcal{D}_e\left(\bm{p}_i,\bm{p}_k\right)=2\tanh(\|\bm{p}_i -\bm{p}_k\|_2)$. $\lambda\in [0, 1]$ is a hyperparameter controlling the contribution of feature and coordinate information.  Similarly, the elements of $\bm{C}^q\in\mathbb{R}^{M\times M}$ related to $\{\bm{q}_j, \bm{q}_l\}$ for $\bm{\mathcal{Q}}$ satisfy
\begin{equation}\label{eq:mq}
	\bm{C}^q_{jl}=\lambda\mathcal{D}_e\left(\bm{q}_j,\bm{q}_l\right)+(1-\lambda)\mathcal{D}_f\left(\bm{f}_{\bm{q}_j},\bm{f}_{\bm{q}_l}\right).
\end{equation} 

We jointly exploit the pointwise and structural matchings equipped with overlap scores to indicate the correspondences between source and target point clouds. This leads to an optimization problem that can be solved by a coupled optimal transport method (called coupled optimal transport since it contains two types of distance, i.e., WD and GWD).
\begin{equation}\label{eq:fot}
	\begin{aligned}
		&\min_{\Gamma}\sum_{ij}\xi_1\underbrace{\Gamma_{ij}\bm{C}^{pq}_{ij}}_{\mbox{WD}} + \xi_2\sum_{ijkl}\underbrace{\Gamma_{ij}\Gamma_{k, l}\left(\bm{C}^p_{ik}-\bm{C}^q_{jl}\right)^2}_{\mbox{GWD}},\\
		&\mbox{s.t.,~} \Gamma\bm{1}_M  = \bm{\mu}_p, \Gamma^\top\bm{1}_N = \bm{\mu}_q, \Gamma_{ij}\in[0, 1],
	\end{aligned}
\end{equation}
where $\bm{1}_n$ denotes an $n$-dimensional all-one vector. $\xi_1$ and $\xi_2$ are two non-negative hyperparameters controlling the pointwise and structural matchings, respectively. If $\xi_1>0$ and $\xi_2=0$, then it only depends on the pointwise matching, and if $\xi_1=0$ and $\xi_2>0$, it only considers the structural matching.  When $\xi_1>0$ and $\xi_2>0$, it allows the framework to effectively consider both pointwise and structural matchings for better correspondence predictions by sharing the assignment matrix $\Gamma$.  After obtaining $\Gamma$, we can apply some methods, such as SVD and RANSAC, to estimate the transformation.

Next, we introduce the Wasserstein distance-based pointwise matching and the Gromov-Wasserstein
distance-based structural matching, respectively.

\subsection{Wasserstein Distance-Based Pointwise Matching} \label{subs:point}
Our Wasserstein distance-based pointwise matching method can be regarded as a variant of the nearest neighbor search with an additional bijectivity constraint that enforces global matching consistency. Our model is appropriate for partial registration, since the overlap scores have been introduced. Given two point clouds $\bm{\mathcal{P}}$ and $\bm{\mathcal{Q}}$, with their associated features $\bm{\mathcal{F}}_p$ and $\bm{\mathcal{F}}_q$, overlap scores $\bm{\mu}_p$ and $\bm{\mu}_q$, the assignment matrix $\Gamma$ can be estimated based on the following Theorem.
\begin{theorem}\label{thm:point}
	Given features $\bm{\mathcal{F}}_p$ and $\bm{\mathcal{F}}_q$, overlap scores  $\bm{\mu}_p$ and $\bm{\mu}_q$,	if $\bm{\mathcal{F}}_p, \bm{\mathcal{F}}_q$ are invariant to rigid transformations, and $\bm{R}^\star, \bm{t}^\star, \Gamma^\star$ are the optimal solutions of problem in Eq. \eqref{eq:reg0}. Then $\Gamma^\star$ is an optimal solution to the following optimization problem:
	\begin{equation}\label{eq:point}
		\begin{aligned}
			& \min_{\Gamma}\left<\bm{C}^{pq}, \Gamma\right> = \min_{\Gamma}\sum_{i=1}^N\sum_{j=1}^M\Gamma_{ij}\bm{C}^{pq}_{ij},\\
			& \mbox{s.t.}~ \Gamma\bm{1}_M  = \bm{\mu}_p, \Gamma^\top\bm{1}_N = \bm{\mu}_q, \Gamma_{ij}\in[0, 1],
		\end{aligned}
	\end{equation}
	where $\bm{f}_{\bm{p}_i}\in \bm{\mathcal{F}}_p$, $\bm{f}_{\bm{q}_j}\in \bm{\mathcal{F}}_q$ represent the features of points $\bm{p}_i$ and $\bm{q}_j$, respectively. $\bm{C}^{pq}_{ij}=\|\frac{\bm{f}_{\bm{p}_i}}{\|\bm{f}_{\bm{p}_i}\|_2}-\frac{\bm{f}_{\bm{q}_j}}{\|\bm{f}_{\bm{q}_j}\|_2}\|_2$. The constraint of the assignment matrix $\Gamma$ is relaxed to a doubly stochastic state, that is, $ \Gamma_{ij}\in[0, 1]$. 
\end{theorem}

Please refer to the proof of this theorem in Appendix \ref{ap:thm1}. 
\begin{remark}
Theorem \ref{thm:point} signifies that the assignment matrix $\Gamma$ can be calculated by solving the optimization problem in Eq. \eqref{eq:point}, which is related to an optimal transport \cite{peyre2019computational} problem (Wasserstein distance). 
It can be solved using the Sinkhorn algorithm \cite{cuturi2013sinkhorn}.   
\end{remark}

\subsection{Gromov-Wasserstein Distance-Based Structural Matching}\label{subs:line}
Our method is based on the observations as follows: for $\forall ~ \bm{p}_i, \bm{p}_k\in\bm{\mathcal{P}}$ and $\forall~\bm{q}_j, \bm{q}_l\in\bm{\mathcal{Q}}$ with their associated features $\bm{f}_{\bm{p}_i}, \bm{f}_{\bm{p}_k}\in\bm{\mathcal{F}}_p$ and $\bm{f}_{\bm{q}_j}, \bm{f}_{\bm{q}_l}\in\bm{\mathcal{F}}_q$, if $\bm{p}_i\rightarrow\bm{q}_j$ and $\bm{p}_k\rightarrow\bm{q}_l$ are correct correspondences, then the distance between $\bm{p}_i$ and $ \bm{p}_k$ should be similar to the distance between $\bm{q}_j$ and $\bm{q}_l$. Thus, the structural difference in both Euclidean and feature spaces should be small, i.e., $|c^e\left(\bm{p}_i,\bm{p}_k\right)-c^e\left(\bm{q}_j,\bm{q}_l\right)|$ and $|c^f\left(\bm{f}_{\bm{p}_i},\bm{f}_{\bm{p}_k}\right)-c^f\left(\bm{f}_{\bm{q}_j}, \bm{f}_{\bm{q}_l}\right)|$
should be small. Gromov-Wasserstein distance is often applied to find the correspondences between two sets of samples based on their pairwise intra-domain similarity (or distance) matrices. Thus, we can approximately transform the correspondence prediction to a structural matching problem based on the Gromov-Wasserstein distance as the following Theorem.
\begin{theorem}\label{thm:line}
	Given two point clouds $\bm{\mathcal{P}}$ and $\bm{\mathcal{Q}}$, with their associated features $\bm{\mathcal{F}}_p$ and $\bm{\mathcal{F}}_q$, overlap scores $\bm{\mu}_p$ and $\bm{\mu}_q$, if $\bm{\mathcal{F}}_p, \bm{\mathcal{F}}_q$ are invariant to rigid transformation, and $\bm{R}^\star, \bm{t}^\star, \Gamma^\star$ are the optimal solutions of problem (\ref{eq:reg0}), then $\Gamma^\star$ is an optimal solution of the following Gromov-Wasserstein distance-based optimization:
	\begin{equation}\label{eq:structure}
		\begin{aligned}
			& \min_{\Gamma}\sum_{i=1}^N\sum_{j=1}^M\sum_{k=1}^{N}\sum_{l=1}^{M}\Gamma_{ij}\Gamma_{kl}\left(\bm{C}^p_{ik}-\bm{C}^q_{jl}\right)^2 \\
			& \mbox{s.t.,~} \Gamma\bm{1}_M  = \bm{\mu}_p, \Gamma^\top\bm{1}_N = \bm{\mu}_q, \Gamma_{ij}\in[0, 1],
		\end{aligned}
	\end{equation}
	where $\bm{C}^p_{ik}$ and $\bm{C}^q_{jl}$ are defined as Eq. \eqref{eq:mp} and Eq. \eqref{eq:mq}, respectively. 
\end{theorem}
The proof is available in Appendix \ref{ap:thm2}.

\begin{remark}
	Assignment matrix $\Gamma$ can be obtained from by problem in Eq. \eqref{eq:structure} by solving an entropy regularized optimization. The structural	matching is formulated by jointly considering the structural differences in both Euclidean and feature space. 
\end{remark}

\subsection{Model Optimization}
Now, we introduce how to solve the problem in \eqref{eq:fot}. For simplicity, we denote a matrix $\bm{H}\left(\bm{C}^p, \bm{C}^q, \Gamma\right)\in\mathbb{R}^{N\times M}$ with each element $[\bm{H}\left(\bm{C}^p, \bm{C}^q, \Gamma\right)]_{kl}=\sum_{i=1}^N\sum_{j=1}^{M}\left(\bm{C}^p_{ik}- \bm{C}^q_{jl}\right)^2\Gamma_{ij}$.  We get
$\sum_{ijkl}\Gamma_{ij}\Gamma_{kl}\left(\bm{C}^p_{ik}- \bm{C}^q_{jl}\right)^2=\left<\bm{H}\left(\bm{C}^p, \bm{C}^q, \Gamma\right), \Gamma\right>$.
The problem in Eq. \eqref{eq:fot} can be rewritten as
\begin{subequations}
	\begin{alignat}{2}
		&\min_{\Gamma\geq 0}\xi_1\underbrace{\left<\bm{C}^{pq}, \Gamma\right>}_{\mbox{WD}} + \xi_2\underbrace{\left<\bm{H}\left(\bm{C}^p, \bm{C}^q, \Gamma\right), \Gamma\right>}_{\mbox{GWD}}, \label{eq:obj}\\
		&\mbox{s.t.,~} \Gamma\bm{1}_M  = \bm{\mu}_p, \Gamma^\top\bm{1}_N = \bm{\mu}_q. \label{eq:cons}
	\end{alignat}
	\label{eq:matrix}
\end{subequations}
%\begin{equation*}
%	\begin{aligned}
	%		g\left(\Gamma\right) & := \xi_1\left<\bm{C}^{pq}, \Gamma\right> + \xi_2\left<\bm{H}\left(\bm{C}^p, \bm{C}^q, \Gamma\right), \Gamma\right>\\
	%		& = \left<\xi_1\bm{C}^{pq} + \xi_2\bm{H}\left(\bm{C}^p, \bm{C}^q, \Gamma\right),  \Gamma\right>,
	%	\end{aligned}
%\end{equation*}

Standard optimal transport only allows a meaningful comparison of measures with the same total mass, i.e., $\sum_{i=1}^{N}\bm{\mu}_{p_i}=\sum_{j=1}^{M}\bm{\mu}_{q_j}$, which does not always satisfy the registration requirement due to multiple correspondences. Following \cite{pham2020unbalanced}, we replace the constraints in Eq. \eqref{eq:cons} with soft-marginals ($\mathcal{KL}$ divergence). Optimization in Eq. \eqref{eq:matrix} is then translated into an unconstrained approximate transport problem
\begin{equation}\label{eq:rotp}
	\begin{aligned}
		\min_{\Gamma\geq 0} & \xi_1\left<\bm{C}^{pq}, \Gamma\right> + \xi_2\left<\bm{H}\left(\bm{C}^p, \bm{C}^q, \Gamma\right), \Gamma\right> \\
		& + \tau\left(\mathcal{KL}\left(\Gamma\bm{1}_M |\bm{\mu}_p\right) + \mathcal{KL}\left(\Gamma^\top\bm{1}_N |\bm{\mu}_q\right)\right),
	\end{aligned}
	%	 g\left(\Gamma\right) + \tau\left(\mathcal{KL}\left(\Gamma\bm{1}_M |\bm{\mu}_p\right) + \mathcal{KL}\left(\Gamma^\top\bm{1}_N |\bm{\mu}_q\right)\right),
\end{equation}
where $\tau > 0$ is a regularization parameter to adjust the strength of penalization of the soft margins. We use the generalized proximal point method \cite{iusem2000dual} and projected gradient descent to solve the problem in Eq. \eqref{eq:rotp} based on the $\mathcal{KL}$ metric. Following \cite{solomon2016entropic,iusem2000dual}, we fix $\Gamma^{(k)}$ at iteration $ k + 1 $ for $k\geq 0$, $\mathcal{KL}\left(\Gamma|\Gamma^{(k)}\right)$ acts as a regularization centered on the previous solution $\Gamma^{(k)}$. The update rule for Eq. \eqref{eq:rotp} at iteration $ k + 1 $ can be written as
\begin{equation}\label{eq:update}
	\begin{aligned}
		\Gamma^{(k+1)} & = \mathop{\arg\min}_{\Gamma\geq 0} \epsilon \mathcal{KL}\left(\Gamma|\Gamma^{(k)}\right) \\
		& +\xi_1\left<\bm{C}^{pq}, \Gamma\right> + \xi_2\left<\bm{H}\left(\bm{C}^p, \bm{C}^q, \Gamma^{(k)}\right), \Gamma\right> \\
		&  + \tau\left(\mathcal{KL}\left(\Gamma\bm{1}_M |\bm{\mu}_p\right) + \mathcal{KL}\left(\Gamma^\top\bm{1}_N |\bm{\mu}_q\right)\right),
	\end{aligned}
\end{equation}
with initialization $\Gamma^{(0)}=\bm{\mu}_p{\bm{\mu}_q}^\top$. $\epsilon > 0$ is a regularization parameter. $\mathcal{KL}\left(\Gamma|\Gamma^{(k)}\right) $ can be interpreted as a damping term that encourages $\Gamma^{(k+1)}$ not to be very far from $\Gamma^{(k)}$. For $\epsilon$ small enough, $\Gamma^{(k+1)}$ in Eq. \eqref{eq:update} converges to the optimal solution of problem in Eq. \eqref{eq:matrix} as $\tau$ increases. Choosing $\epsilon$ trades off convergence speed with closeness to the original transport problem \cite{cuturi2013sinkhorn}. The solution of the problem in Eq. \eqref{eq:update} is based on the following theorem. 

\begin{theorem}\label{thm:dual}
	Denote $f(\Gamma^{(k)})=\xi_1\bm{C}^{pq} +\xi_2\bm{H}\left(\bm{C}^p, \bm{C}^q, \Gamma^{(k)}\right) - \epsilon\log \Gamma^{(k)}$ and $\bm{C}\in\mathbb{R}^{N\times M}$ with elements that satisfy $\bm{C}_{ij}=[f\left(\Gamma^{(k)}\right)]_{ij}$. The optimal solution for the objective in Eq. \eqref{eq:iter11} can be obtained by solving the following dual entropic regularized objective,
	\begin{equation}\label{eq:dual3}
		\begin{aligned}
			& \min_{\bm{u},\bm{v}} h(\bm{u}, \bm{v}) = \min_{\bm{u},\bm{v}}\epsilon \sum_{i=1}^N\sum_{j=1}^{M}\exp\left(\frac{\bm{u}_i+\bm{v}_j-\bm{C}_{ij}}{\epsilon}\right) \\ 
			& + \tau\left\langle\exp\left(-\frac{\bm{u}}{\tau}\right),\bm{\mu}_p\right\rangle + \tau\left\langle\exp\left(-\frac{\bm{v}}{\tau}\right),\bm{\mu}_q\right\rangle,
		\end{aligned}
	\end{equation}
	where $\bm{u}\in\mathbb{R}^N,\bm{v}\in\mathbb{R}^M$ are dual variables. 
\end{theorem}

Please refer to the proof of the theorem in Appendix \ref{ap:thm3}. The problem in Eq. \eqref{eq:dual3} can be solved using the Sinkhorn algorithm \cite{cuturi2013sinkhorn,pham2020unbalanced}. Specifically,
\begin{equation*}
	\begin{aligned}
		& \frac{\partial h}{\partial\bm{u}} = 0 \Rightarrow \sum^N_j\exp\left(\frac{\bm{u}_i+\bm{v}_j-\bm{C}_{ij}}{\epsilon}\right) - \exp\left(-\frac{\bm{u}_i}{\tau}\right)\bm{\mu}_{\bm{p}_i} = 0\\
		& \Rightarrow \exp\left(\frac{\bm{u}_i}{\epsilon}\right)\sum^N_j\exp\left(\frac{\bm{v}_j-\bm{C}_{ij}}{\epsilon}\right) = \exp\left(-\frac{\bm{u}_i}{\tau}\right)\bm{\mu}_{\bm{p}_i}.	
	\end{aligned}
\end{equation*}
Let $(\bm{u}^k, \bm{v}^k, \bm{a}^k, \bm{b}^k)$ be the solution returned at the $k$-th iteration of the algorithm. Here $\bm{a}=B(\bm{u}, \bm{v})\textbf{1}_M$ and $\bm{b}=B(\bm{u}, \bm{v})^\top\textbf{1}_N$ with $B(\bm{u}, \bm{v})= \text{diag}\left(\exp{\left(\frac{\bm{u}}{\epsilon}\right)}\right)\cdot \exp{\left(-\frac{\bm{C}}{\epsilon}\right)}\cdot\text{diag}\left(\exp{\left(\frac{\bm{v}}{\epsilon}\right)}\right)$. Suppose we are at iteration $ k + 1 $ for $k\geq 0$ with a fixed $\bm{v}^k$, i.e., 
\begin{equation*}
	\exp\left(\frac{\bm{u}_i^{k+1}}{\epsilon}\right)\sum^N_j\exp\left(\frac{\bm{v}_j^{k}-\bm{C}_{ij}}{\epsilon}\right) = \exp\left(-\frac{\bm{u}_i^{k+1}}{\tau}\right)\bm{\mu}_{\bm{p}_i}.
\end{equation*} 
Multiplying both sides by $\exp\left(\frac{\bm{u}_i^{k}}{\epsilon}\right)$, we get 
\begin{equation*}
	\begin{aligned}
		&\exp\left(\frac{\bm{u}_i^{k+1}}{\epsilon}\right)\bm{a}_i^k = \exp\left(\frac{\bm{u}_i^{k}}{\epsilon}\right)\exp\left(-\frac{\bm{u}_i^{k+1}}{\tau}\right)\bm{\mu}_{\bm{p}_i} \\
		& \Rightarrow \bm{u}^{k+1}=\left[\frac{\bm{u}^{k}}{\epsilon} + \log\left(\bm{\mu}_p\right)-\log\left(\bm{a}^k\right)\right]\frac{\epsilon\tau}{\epsilon+\tau}.
	\end{aligned}
\end{equation*}
Similarly, with $\bm{u}^k$ fixed, we get
\begin{equation*}
	%	\exp\left(\frac{\bm{v}_j^{k+1}}{\epsilon}\right)\bm{b}_j^k = \exp\left(\frac{\bm{v}_j^{k}}{\epsilon}\right)\exp\left(-\frac{\bm{v}_j^{k+1}}{\tau}\right)\bm{\mu}_{\bm{q}_i}.
	\bm{v}^{k+1}=\left[\frac{\bm{v}^{k}}{\epsilon} + \log\left(\bm{\mu}_q\right)-\log\left(\bm{b}^k\right)\right]\frac{\epsilon\tau}{\epsilon+\tau}.
\end{equation*}
We use the pseudocode in Algorithm \ref{alg:alg1} to illustrate the solution. The inner iterations can be determined by $\epsilon, \bm{C}_{ij}$ and $\max\{M, N\}$ and the proof is similar to Theorem \ref{thm:line} in \cite{pham2020unbalanced}. In our experiments, we found that when setting $\xi_1=1.0$, $\tau=5.0$, $\epsilon=0.001$, $N_I = 100$ and $N_O=20$, it can obtain satisfactory results.

\begin{algorithm}[t]
	\caption{Coupled optimal transport algorithm.}  
	\label{alg:alg1} 
	\hspace*{0.02in} {\bf Input:} Distance matrices $\bm{C}^p$, $\bm{C}^q$ and $\bm{C}^{pq}$, overlap scores $\bm{\mu}_p$ and $\bm{\mu}_q$, and hyparameters  $\tau,\epsilon>0$, $\xi_1=1$, and the number of outer/inner iterations $N_O, N_I$.
	\begin{algorithmic}[1]
		\STATE Initialize $\Gamma^{(0)}=\bm{\mu}_p{\bm{\mu}_q}^\top$, $k=0$.
		\FOR{$k = 0 : N_O$}
		\STATE $\xi_2=\frac{k}{N_O}$
		\STATE Compute $\bm{C}_{ij} = \left[f\left(\Gamma^{(k)}\right)\right]_{ij}$
		\WHILE{$k < N_I$}
		\STATE $\bm{a}^k=B(\bm{u}^k, \bm{v}^k)\textbf{1}_M, \bm{b}^k=B(\bm{u}^k, \bm{v}^k)^\top\textbf{1}_N$.
		\IF{$k$ is even}
		\STATE $\bm{u}^{k+1}=\left[\frac{\bm{u}^{k}}{\epsilon} + \log\left(\bm{\mu}_p\right)-\log\left(\bm{a}^k\right)\right]\frac{\epsilon\tau}{\epsilon+\tau}$
		\STATE $\bm{v}^{k+1}=\bm{v}^{k}$
		\ELSE 
		\STATE $\bm{v}^{k+1}=\left[\frac{\bm{v}^{k}}{\epsilon} + \log\left(\bm{\mu}_q\right)-\log\left(\bm{b}^k\right)\right]\frac{\epsilon\tau}{\epsilon+\tau}$
		\STATE $\bm{u}^{k+1}=\bm{u}^{k}$
		\ENDIF
		\STATE $k=k+1$.
		\ENDWHILE
		\STATE $\Gamma^{(k)}=B(\bm{u}^{k}, \bm{v}^k)$
		\ENDFOR
		\STATE {\bf Output:} $\Gamma^{(N_O)}$
	\end{algorithmic}
\end{algorithm}

\subsection{Combined with Learning Network}
\begin{figure}[t]
	\centering  
	\includegraphics[width=3.3in]{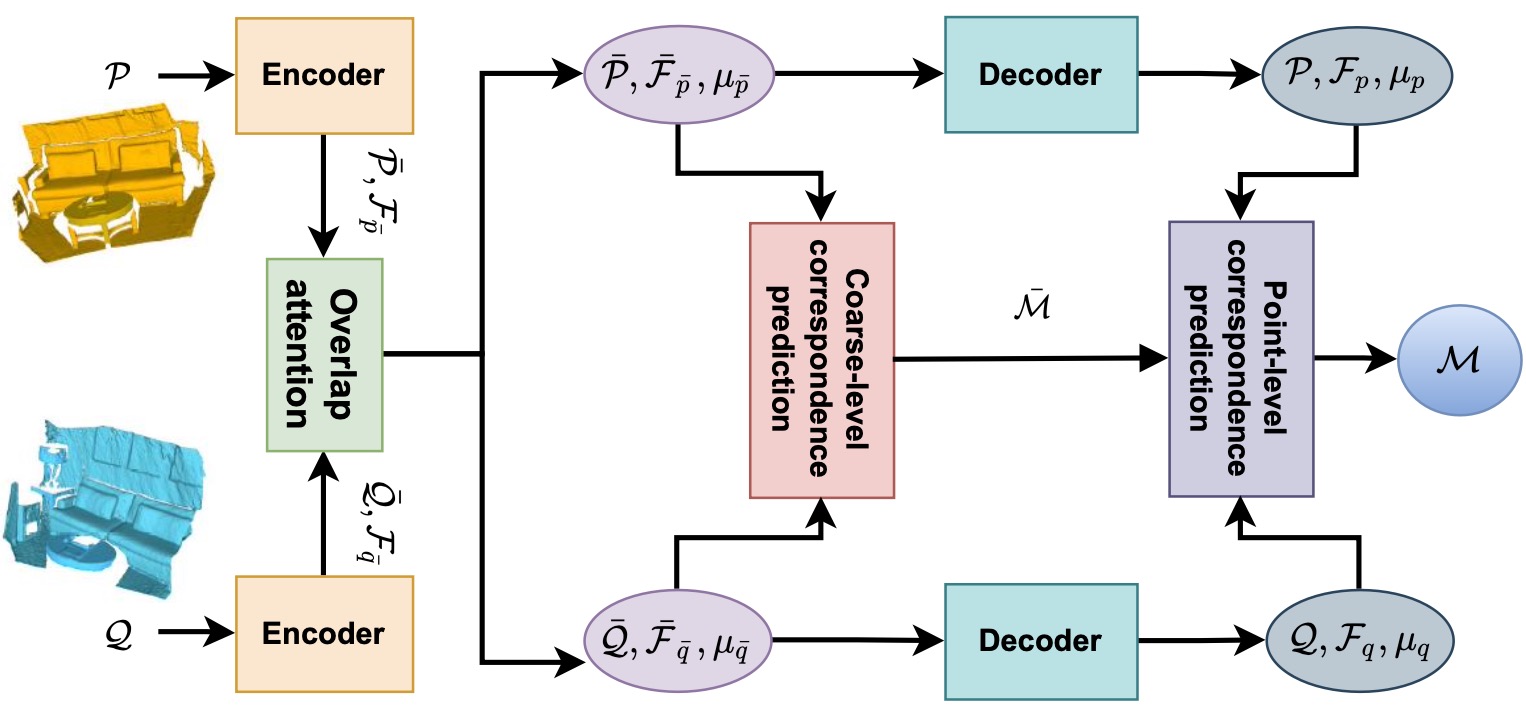}
	\caption{Overview of the proposed COTReg combined with network. COTReg adopts a hierarchical matching strategy that first establishes superpoint-level correspondences and then predicts point-level correspondences according to superpoint-level matches.}
	\label{fig:frame}
\end{figure}

% \review{The proposed pipeline does not use inlier scores in test time.}
% \gmei{Guofeng: The reviewer misunderstand our method, the overlap scores are used in both training and inference stages. Based on Prof. Mohammed suggestion, I give a process to show the detail \textbf{in this whole section}.}
The solution of the optimization problem in Eq.~\eqref{eq:fot} is sought over the space of $N\times M$ permutation matrices. Because of memory constraints and speed limitations, it is not suitable to solve large scale registration problem. To this end, COTReg adopts a hierarchical matching strategy that first establishes superpoint-level correspondences and then predicts point-level correspondences according to superpoint-level matches. The pipeline of our COTReg is illustrated in Figure \ref{fig:frame}, which is a shared weighted two-stream encoder-decoder network. Given a pair of point cloud $\bm{\mathcal{P}}$ and $\bm{\mathcal{Q}}$, the encoder aggregates the raw points into superpoints $\bar{\bm{\mathcal{P}}}=\{\bar{\bm{p}}_i \in\mathbb{R}^{3}|i = 1, 2, ..., \bar{N}\}$ and $\bar{\bm{\mathcal{Q}}}=\{\bar{\bm{q}}_j \in\mathbb{R}^{3}|j = 1, 2, ..., \bar{M}\}$, while jointly learning the associated features $\bm{\mathcal{F}}_{\bar{p}}=\{\bm{f}_{\bar{p}_i} \in\mathbb{R}^{b}|i = 1, 2, ..., \bar{N}\}$ and $\bm{\mathcal{F}}_{\bar{q}}=\{\bm{f}_{\bar{q}_j} \in\mathbb{R}^{b}|j = 1, 2, ..., \bar{M}\}$. The overlap attention block updates the features as $\bar{\bm{\mathcal{F}}}_{\bar{p}}$ and $\bar{\bm{\mathcal{F}}}_{\bar{q}}$, and projects them to coarse level overlap score vectors $\bm{\mu}_{\bar{p}}=\{\bm{\mu}_{\bar{p}_i}\in\left[0,1\right]\}_{i=1}^{\bar{N}}, \bm{\mu}_{\bar{q}}=\{\bm{\mu}_{\bar{q}_j}\in\left[0,1\right]\}_{i=1}^{\bar{M}}$. The updated features and overlap scores are then used to calculate the coarse-level correspondences. Finally, the decoder transforms the superpoint level features and overlap scores to per-point feature descriptors $\bm{\mathcal{F}}_{p}$ and $\bm{\mathcal{F}}_{q}$, and overlap scores $\bm{\mu}_{p}$ and $\bm{\mu}_{q}$, which are used to estimate fine-level correspondences.  
\subsubsection{Encoder}
Inspired by Predator \cite{huang2021predator}, a shared KPConv \cite{thomas2019kpconv}, which consists of a series of ResNet-like blocks and stridden convolutions, simultaneously down-samples the raw point clouds $\bm{\mathcal{P}}$ and $\bm{\mathcal{Q}}$ into superpoints $\bar{\bm{\mathcal{P}}}$ and $\bar{\bm{\mathcal{Q}}}$ and extracts associated features $\bm{\mathcal{F}}_{\bar{p}}$ and $\bm{\mathcal{F}}_{\bar{q}}=\{\bm{f}_{\bar{q}_j} \in\mathbb{R}^{b}|j = 1, 2, ..., \bar{M}\}$, respectively. 

\subsubsection{Overlap Attention Module}
The overlap attention module, which estimates the probability (overlap score) of whether a point is in the overlapping area, consists of positional encoding, self-attention, cross-attention, and overlap score prediction. The positional encoding assigns intrinsic geometric properties to a per-point feature, thus enhancing distinctions among point features in indistinctive regions. The extracted local features have a limited receptive field which may not distinguish indistinctive regions. Instead, humans find correspondences in these indistinctive regions not only based on the local neighborhood but with a larger global context. Self-attention is thus introduced to model the long-range dependencies. And cross attention module exploits the intra-relationship within the source and target point clouds, which models the potential overlap regions. Here, we detail the individual parts hereafter.

\noindent\textbf{Positional Encoding.} 
Positional encoding scheme which assigns intrinsic geometric properties to the per-point feature by adding unique positional information enhances distinctions among point features in indistinctive regions. Given two superpoints $\bar{\bm{p}}_i$ and $\bar{\bm{p}}_j$ of $\bar{\bm{\mathcal{P}}}$, we first select the $k=5$ nearest neighbors $\mathcal{K}_i$ of $\bar{\bm{p}}_i$ and compute the centroid $\bar{\bm{p}}_c=\sum_{i=1}^{\bar{N}}\bar{\bm{p}}_i$ of $\bar{\bm{\mathcal{P}}}$. For each $\bar{\bm{p}}_x\in \mathcal{K}_i$, we denote the angle between two vectors $\bar{\bm{p}}_i-\bar{\bm{p}}_c $ and $ \bar{\bm{p}}_x-\bar{\bm{p}}_c$ is $\alpha_{ix}$.
The position encoding $\bm{f}^{pos}_{\bar{p}_i}$ of $\bar{\bm{p}}_i$ is defined as follows: 
\begin{equation}\label{eq:pos}
	\bm{f}^{pos}_{\bar{p}_i} = \varphi\left(\|\bar{\bm{p}}_i-\bar{\bm{p}}_c\|_2\right) +\max_{x\in \mathcal{K}_i}\{\phi\left(\alpha_{ix}\right)\},
\end{equation}
where $\varphi$ and $\phi$ are two MLPs, and each MLP consists of a linear layer and one ReLU nonlinearity function.

\noindent\textbf{Self-Attention and Cross-Attention.} Let $\bm{\mathcal{F}}^l_{\bar{p}}$ be the intermediate representation for $\bar{\bm{\mathcal{P}}}$ at layer $l$ and let $\bm{\mathcal{F}}^0_{\bar{p}}=\{\bm{f}^{pos}_{\bar{p}_i}+\bm{f}_{\bar{p}_i}\}_{i=1}^{\bar{N}}$. We use a multi-attention layer consisting of
four parallel attention head to update the $\bm{\mathcal{F}}^l_{\bar{p}}$ via
\begin{equation}\label{eq:sim}
	\begin{aligned}
			\bm{S}_{\bar{p}} &= \bm{W}^{l}_1\bm{\mathcal{F}}^l_{\bar{p}} + \bm{b}^{l}_1, \bm{K}_{\bar{x}} = \bm{W}^{l}_2 \bm{\mathcal{F}}^l_{\bar{x}} + \bm{b}^{l}_2, \\
			\bm{V}_{\bar{x}} &= \bm{W}^{l}_3\bm{\mathcal{F}}_{\bar{x}} + \bm{b}^{l}_3, \bm{A} = \mbox{softmax}\left(\bm{S}_{\bar{p}}^\top \bm{K}_{\bar{x}}\big/\sqrt{b}\right), \\
			\bm{\mathcal{F}}^{l+1}_{\bar{p}} &=\bm{\mathcal{F}}_{\bar{p}} + g^{l}\left(\left[\bm{\mathcal{F}}^l_{\bar{p}}\big\|\bm{A}\bm{V}_{\bar{x}}\right]\right).
		\end{aligned}
\end{equation}
Here, if $\bar{x}=\bar{p}$ represents self-attention, and if $\bar{x}=\bar{q}$ indicates cross-attention. $[\cdot\|\cdot]$ denotes concatenation, and $g^{l}\left(\cdot\right)$ is a three-layer fully connected network consisting of a linear layer, instance normalization, and a LeakyReLU activation. The same attention module is also simultaneously performed for all points in point cloud $\bar{\bm{\mathcal{Q}}}$. A fixed number of layers $L=2$ with different parameters are chained and alternatively aggregate along the self- and cross-attention. As such, start from $l = 0, \bar{x}=\bar{p}$ if $l$ is even and $\bar{x}=\bar{q}$ if $l$ is odd. The final outputs of attention module are $\bar{\bm{\mathcal{F}}}_{\bar{p}}=\bm{\mathcal{F}}^{3}_p$ for $\bar{\bm{\mathcal{P}}}$ and $\bar{\bm{\mathcal{F}}}_{\bar{q}}=\bm{\mathcal{F}}^3_{\bar{q}}$ for $\bar{\bm{\mathcal{Q}}}$. By doing this, each point can incorporate non-local information that intuitively strengthens their long-range correlation dependencies. The latent features $\bar{\bm{\mathcal{F}}}_{\bar{p}}$ have the knowledge of $\bar{\bm{\mathcal{F}}}_{\bar{q}}$ and vice versa. 

\noindent\textbf{Overlap Score Prediction.}
To deal with those points in non-overlapping regions, we separately predict the overlap scores $\bm{\mu}_{\bar{p}}=\{\bm{\mu}_{\bar{p}_1},  \bm{\mu}_{\bar{p}_2}, \cdots, \bm{\mu}_{\bar{p}_{\bar{N}}}\}$ and $\bm{\mu}_{\bar{q}}=\{\bm{\mu}_{\bar{q}_1},  \bm{\mu}_{\bar{q}_2}, \cdots, \bm{\mu}_{\bar{q}_{\bar{M}}}\}$ using the conditioned features $\bar{\bm{\mathcal{F}}}_{\bar{p}}$ and $\bar{\bm{\mathcal{F}}}_{\bar{q}}$. Here $\bm{\mu}_{\bar{p}_i} \in [0, 1]$ and $\bm{\mu}_{\bar{q}_j} \in [0, 1]$ can be computed using a single fully layer $g_\beta\left(\cdot\right)$ followed by a sigmoid activation function.
\begin{equation}\label{eq:opcom}
\begin{aligned}
\bm{\mu}_{\bar{p}} &= \mbox{sigmoid}\left(g_\beta\left(\bar{\bm{\mathcal{F}}}_{\bar{p}}\right)\right), \\
	\bm{\mu}_{\bar{q}} &= \mbox{sigmoid}\left(g_\beta\left(\bar{\bm{\mathcal{F}}}_{\bar{q}}\right)\right).
\end{aligned}
\end{equation}
The overlap scores are used to mask out the influence of points outside the overlap region.

\subsubsection{Coarse-Level Correspondence Prediction} 
$\bar{\bm{\mathcal{F}}}_{\bar{p}}$, $\bar{\bm{\mathcal{F}}}_{\bar{q}}$, $\bm{\mu}_{\bar{p}}$ and $\bm{\mu}_{\bar{q}}$ are used to calculate an assignment matrix $\bar{\bm{\Gamma}}$, at superpoint level, by solving a coupled optimal transport problem
\begin{equation}\label{eq:coarse}
	\begin{aligned}
		\min_{\bar{\bm{\Gamma}}\geq 0} & \left<\xi_1\bm{C}^{\bar{p} \bar{q}}, \bar{\bm{\Gamma}}\right> + \left<\xi_2\bm{H}\left(\bm{C}^{\bar{p}}, \bm{C}^{\bar{q}}, \bar{\bm{\Gamma}}\right), \bar{\bm{\Gamma}}\right> \\
		& + \tau\left(\mathcal{KL}\left(\bar{\bm{\Gamma}}\bm{1}_{\bar{M}} |\bm{\mu}_{\bar{p}}\right) + \mathcal{KL}\left(\bar{\bm{\Gamma}}^\top\bm{1}_{\bar{N}} |\bm{\mu}_{\bar{q}}\right)\right),
	\end{aligned}
\end{equation}
where ${\bm{C}}^{\bar{p} \bar{q}},\bm{C}^{\bar{p}}$, and $\bm{C}^{\bar{q}}$ are the distance matrices with elements satisfying ${\bm{C}}^{\bar{p} \bar{q}}_{ij}=\mathcal{D}_f\left(\bar{\bm{f}}_{\bar{p}_i},\bar{\bm{f}}_{\bar{q}_j}\right),\bm{C}^{\bar{p}}_{ij}=\mathcal\lambda{D}_e\left(\bar{p}_i,\bar{p}_j\right)+(1-\lambda)\mathcal{D}_f\left(\bar{\bm{f}}_{\bar{p}_i},\bar{\bm{f}}_{\bar{p}_j}\right)$ and ${\bm{C}}^{\bar{q} }_{ij}=\mathcal\lambda{D}_e\left(\bar{q}_i,\bar{q}_j\right)+(1-\lambda)\mathcal{D}_f\left(\bar{\bm{f}}_{\bar{q}_i},\bar{\bm{f}}_{\bar{q}_j}\right)$, respectively. $\lambda>0$ is a hyperparameter. Eq. \eqref{eq:coarse} is an instance of the optimal transport \cite{cuturi2013sinkhorn} problem, which can be solved efficiently using Sinkhorn-Knopp algorithm \cite{cuturi2013sinkhorn}. After reaching $\bar{\bm{\Gamma}}$, we select correspondences with maximum confidence score of $\bar{\bm{\Gamma}}$ in each row and column, and further enforce the mutual nearest neighbor (MNN) criterion, which filters possible outlier coarse matches. The coarse-level correspondences are defined as follows:
\begin{equation}\label{eq:sset}
	\bar{\mathcal{M}}=\{(\bar{\bm{p}}_{\hat{i}},\bar{\bm{p}}_{\hat{j}})\big|\forall(\hat{i},\hat{j})\in \mbox{MNN}(\bar{\bm{\Gamma}}), \hat{j} = \arg\max_{k}\bar{\bm{\Gamma}}_{\hat{i}k}\}.
\end{equation}

\subsubsection{Decoder} Our decoder starts with conditioned features $\bm{\mathcal{F}}_{\bar{p}}$, concatenates them with the overlap score $\bm{\mu}_{\bar{p}}$, and outputs the per-point feature descriptor $\bm{\mathcal{F}}_{p}\in\mathbb{R}^{N\times 32}$ and refined per-point overlap scores $\bm{\mu}_{p}$. The decoder architecture combines NN-upsampling with linear layers and includes skip connections from the corresponding encoder layers. The same operator is applied to generate $\bm{\mathcal{F}}_{q}\in\mathbb{R}^{M\times 32}$ and $\bm{\mu}_{q}$.

\subsubsection{Fine-Level Prediction}
On the finer stage, we refine coarse correspondences to point-level correspondences. Those refined matches are then utilized for point cloud registration. 

\noindent\textbf{Point-Level Correspondence Prediction.}
We first group the points into clusters by assigning points to their nearest superpoints in geometry space. After grouping, points with their associated overlap scores and descriptors form patches, on which we can extract point correspondences. For a superpoint $\bar{\bm{p}}_i\in \bar{\bm{\mathcal{P}}}$, its associated point set $G_{\bar{p}_i}$, feature set $G_{\bm{f}_{\bar{p}_i}}$, and the overlap score set $G_{\bm{\mu}_{\bar{p}_i}}$ are denoted as
\begin{equation}\label{eq:group}
	\begin{cases}
		G_{\bar{p}_i}=\{\bm{p}\in \bm{\mathcal{P}}\big|\|\bm{p}-\bm{\bar{p}}_i\|_2\leq\|\bm{p}-\bm{\bar{p}}_j\|_2, i\neq j\}, \\
		G_{\bm{f}_{\bar{p}_i}}=\{\bm{f}_{\bm{x}_j}\in \bm{\mathcal{F}}_{p}\big|\bm{x}_j \in G_{\bar{p}_i}\}, \\
		G_{\bm{\mu}_{\bar{p}_i}}=\{\mu_{\bm{x}_j}\in \bm{\mathcal{\mu}}_{p}\big|\bm{x}_j \in G_{\bar{p}_i}\}.
	\end{cases}
\end{equation}
The coarse-level correspondence set $\bar{\mathcal{M}}$ is expanded to its corresponding patch correspondence sets, both in geometry space $\mathcal{M}_C=\{(G_{\bar{p}_i}, G_{\bar{q}_j})\}$, feature space $\mathcal{M}_F=\{(G_{\bm{f}_{\bar{p}_i}}, G_{\bm{f}_{\bar{q}_j}})\}$, and overlap scores $\mathcal{M}_{\bm{\mu}}=\{(G_{\bm{\mu}_{\bar{p}_i}}, G_{\bm{\mu}_{\bar{q}_j}})\}$. For computational efficiency, every patch samples $K$ number of points based on the overlap scores. Given a pair of overlapped patches $(G_{\bar{p}_i}, G_{\bm{f}_{\bar{p}_i}},G_{\bm{\mu}_{\bar{p}_i}})$ and $(G_{\bar{q}_j}, G_{\bm{f}_{\bar{q}_j}},G_{\bm{\mu}_{\bar{q}_j}})$, we first calculate the cross distance matrix ${\bm{C}}^{\bar{p}_i \bar{q}_j}=\{{\bm{C}}^{\bar{p}_i \bar{q}_j}_{kl}\}$, and structural matrices
$\bm{C}^{\bar{p}_i}=\{\bm{C}^{\bar{p}_i}_{kl}\}$ and $\bm{C}^{\bar{q}_j}=\{\bm{C}^{\bar{q}_j}_{kl}\}$ with elements satisfying 
\begin{equation*}
    \begin{aligned}
        &{\bm{C}}^{\bar{p}_i \bar{q}_j}_{kl}=\mathcal{D}_f\left(G^k_{\bm{f}_{\bar{p}_i}},G^l_{{\bm{f}_{\bar{q}_j}}}\right), \\ 
        &\bm{C}^{\bar{p}_i}_{kl} =\lambda\mathcal{D}_e\left(G^k_{{\bar{p}_i}},G^l_{{\bar{p}_i}}\right) + (1-\lambda)\mathcal{D}_f\left(G^k_{\bm{f}_{\bar{p}_i}},G^l_{\bm{f}_{\bar{p}_i}}\right), \\
        &\bm{C}^{\bar{q}_j}_{kl} =\lambda\mathcal{D}_e\left(G^k_{{\bar{q}_j}},G^l_{{\bar{q}_j}}\right) + (1-\lambda)\mathcal{D}_f\left(G^k_{\bm{f}_{\bar{q}_j}},G^l_{\bm{f}_{\bar{q}_j}}\right),
    \end{aligned}
\end{equation*}
where $\lambda\in[0,1]$ is a hyperparameter.
Extracting point correspondences is analogous to matching two smaller-scale point clouds by solving a coupled optimal transport problem to calculate a matrix $\Gamma_{\bar{p}_i}$ as
\begin{equation*}
		\begin{aligned}
		\min_{\Gamma_{\bar{p}_i}\geq 0} & \left<\xi_1\bm{C}^{\bar{p}_i \bar{q}_j}, \bm{\Gamma}_{\bar{p}_i}\right> + \left<\xi_2\bm{H}\left(\bm{C}^{\bar{p}_i}, \bm{C}^{\bar{q}_j}, {\bm{\Gamma}}_{\bar{p}_i}\right), {\bm{\Gamma}}_{\bar{p}_i}\right> \\
		& + \tau\left(\mathcal{KL}\left(\bm{\Gamma}_{\bar{p}_i}\bm{1}_{M_{\bar{q}_j}} |G_{\bm{\mu}_{\bar{p}_i}}\right) + \mathcal{KL}\left(\bm{\Gamma}_{\bar{p}_i}^\top\bm{1}_{N_{\bar{q}_j}} |G_{\bm{\mu}_{\bar{q}_j}}\right)\right),
	\end{aligned}
\end{equation*}
For correspondences, we choose the maximum confidence score of $\Gamma_{\bar{p}_i}$ in every row and column to guarantee a higher precision. The final point correspondence set $\mathcal{M}$ is represented as the union of all the correspondence sets obtained. After obtaining the correspondences $\mathcal{M}$, following \cite{qin2022geometric,yu2021cofinet},  a variant of RANSAC \cite{fischler1981random} that is specialized to 3D correspondence-based registration \cite{zhou2018open3d} is utilized to estimate the transformation.
% \review{The most serious part is that **the approach estimates 6D transformation using RANSAC and refine the results using ICP**. This is quite misleading because the whole equations and Theorems are devised for estimating “correspondences” and predicting “confidence”, and they are claimed as a main technical contribution. Although the pipeline optimizes to predict valuable information, the approach applies RANSAC and ICP to find the correspondence **again**.}
% {\color{blue} Guofeng: we give an explanation here, and now we do not use the icp in our experiments.}

\subsection{Loss Function and Training}
Our model is an end-to-end learning framework, using the ground truth correspondences as supervision. The loss function $\mathcal{L} = \mathcal{L}_{C}  + \mathcal{L}_{F} + \mathcal{L}_{CO}+ \mathcal{L}_{FO}$ is composed of an coarse-level loss $\mathcal{L}_{C}$ for superpoint matching, a point matching loss $\mathcal{L}_{F}$ for point matching, a binary classification loss $\mathcal{L}_{CO}$ for coarse-level overlap scores, and a classification loss $\mathcal{L}_{FO}$ for fine-level overlap scores.

\subsubsection{Coarse-Level Loss} 
\noindent\textbf{Superpoint Matching Loss.}  Existing methods \cite{yu2021cofinet,fu2021robust} usually formulate superpoint matching as a multilabel classification problem and adopt a cross-entropy loss with optimal transport. Doing this requires unfolding the Sinkhorn layer to compute gradients in the training stage. To address this issue, we adopt a circle loss \cite{sun2020circle} to optimize the superpoint-wise feature descriptors. As there is not direct supervision for superpoint matching, we leverage the overlap ratio $r^j_{i}$ of points in $G_{\bar{p}_i}$ that have correspondences in $G_{\bar{q}_j}$ to depict the matching probability between superpoints $\bar{p}_i$ and $\bar{q}_j$. $r^j_{i}$ is defined as:
\begin{equation*}\label{eq:cratio}
	r^j_{i} =\frac{1}{{|G_{\bar{p}_i}|}} {|\{\bm{p}\in G_{\bar{p}_i}\big|\min_{\bm{q}\in G_{\bar{q}_j}}\|\hat{\bm{T}}\left(\bm{p}\right) -\bm{q}\|_2<r_p\}|}.
\end{equation*}
where $\hat{\bm{T}}$ is the ground-truth transformation and $r_p$ is a set threshold. For circle loss, a pair of superpoints are positive if their corresponded patches share at least $10\%$ overlap, and negative if they do not overlap. All other pairs are omitted. We select the superpoints in $\bm{\bar{\mathcal{P}}}$ which have at least one positive superpoint in $\bm{\bar{\mathcal{Q}}}$ to form a set of anchor superpoints, $\bm{\tilde{\mathcal{P}}}$. For each anchor $\tilde{\bm{p}}_i\in \bm{\tilde{\mathcal{P}}}$,  we denote the set of its positive superpoints in $\bm{\bar{\mathcal{Q}}}$ as $\mathcal{N}_p^{\tilde{\bm{p}}_i}$, and the set of its negative patches as $\mathcal{N}_n^{\tilde{\bm{p}}_i}$. The superpoint matching loss (circle loss) $\mathcal{L}_{C}^{\bm{\bar{\mathcal{P}}}}$ on $\bm{\bar{\mathcal{P}}}$  is then defined as:
\begin{equation}
\begin{aligned}
    \mathcal{L}_{C}^{\bm{\bar{\mathcal{P}}}} &=\frac{1}{|\bm{\tilde{\mathcal{P}}}|} \sum_{\tilde{\bm{p}}_i\in\bm{\bar{\mathcal{P}}}}\log\left[1+\zeta_i\right], \\
    \zeta_i &= \sum_{\bm{\tilde{q}}_k\in\mathcal{N}_p^{\tilde{\bm{p}}_i}}e^{r^k_i\beta_p^{ik}(d_i^k-\Delta p)}\cdot\sum_{\bm{\tilde{q}}_l\in\mathcal{N}_n^{\tilde{\bm{p}}_i}}e^{\beta_n^{il}(\Delta n - d_i^l)},
\end{aligned}
\end{equation}
where $d_i^k=\mathcal{D}_f(\bm{f}_{\tilde{p}_i},\bm{f}_{\tilde{q}_k})$ is the distance in the feature space. The weights $\beta_p^{ik}=\gamma d_i^k$ and $\beta_n^{il}=\gamma (2.0-d_i^l)$ are determined individually for each positive and negative example, using the empirical margins $\Delta p = 0.1$ and $\Delta n = 1.4$ with a learned scale factor $\gamma\geq 1$. The circle loss reweights the loss values on $\mathcal{N}_{p^i}$ based on the overlap ratio so that the patch pairs with higher overlap are given more importance. The same goes for the loss $\mathcal{L}_{C}^{\bm{\bar{\mathcal{Q}}}}$ on $\bm{\bar{\mathcal{Q}}}$. The
overall superpoint matching loss is
\begin{equation}
  \mathcal{L}_{C} = \frac{1}{2}(\mathcal{L}_{C}^{\bm{\bar{\mathcal{P}}}} + \mathcal{L}_{C}^{\bm{\bar{\mathcal{Q}}}}).
\end{equation}

\noindent\textbf{Coarse-Level Overlap Loss.} We use the ratio of points in $G_{\bar{p}_i}$ that are visible in $\mathcal{\bm{Q}}$ to depict the ground-truth overlap scores $\bar{\bm{\mu}}_{\bar{p}_i}$ of the superpoint $\bar{p}_i$. It is calculated by
\begin{equation}\label{eq:gtcop}
	\bar{\bm{\mu}}_{\bar{p}_i} = \frac{1}{{|G_{\bar{p}_i}|}}{|\{\bm{p}\in G_{\bar{p}_i}\big|\min_{\bm{q}\in \mathcal{Q}}\|\hat{\bm{T}}\left(\bm{p}\right) -\bm{q}\|_2<r_o\}|},
\end{equation}
with overlap threshold.
If $\bar{\bm{\mu}}_{\bar{p}_i}$ is close to 1,  $\bar{p}_i$ tends to locate in the overlap regions. $\bar{\bm{\mu}}_{\bar{q}_j}$ is calculated in the same way. The predicted overlap scores for $\bar{\bm{\mathcal{P}}}$ are thus supervised using the binary cross entropy loss, i.e., 
\begin{equation}
   \mathcal{L}_{\bm{\mathcal{\bar{P}}}}  = -\frac{1}{\bar{N}}\sum_{i}\bar{\bm{\mu}}_{\bar{p}_i}\log \bm{\mu}_{\bar{p}_i} + \left(1-\bar{\bm{\mu}}_{\bar{p}_i}\right) \log\left(1-\bm{\mu}_{\bar{p}_i}\right). 
\end{equation}
The loss $\mathcal{L}_{\bm{\bar{\mathcal{Q}}}}$ for $\bm{\bar{\mathcal{Q}}} $
is calculated in the same way. The loss for coarse-level overlap scores is
\begin{equation*}
    \mathcal{L}_{CO} =  \frac{1}{2}\left(\mathcal{L}_{\bar{\bm{\mathcal{P}}}} + \mathcal{L}_{\bar{\bm{\mathcal{Q}}}}\right).
\end{equation*}

\subsubsection{Fine-Level Loss}
\noindent\textbf{Point Matching Loss.}  We apply circle loss again to supervise the point matching. Consider a pair of matched superpoints $\bar{\bm{p}}_i$ and $\bar{\bm{q}}_j$ with associated patches $G_{\bar{p}_i}$ and $G_{\bar{q}_j}$, we first extract a set of anchor points $\tilde{G}_{\bar{p}_i} \subseteq G_{\bar{p}_i}$ satisfying that each $\bm{g}^k_{\bar{p}_i}\in \tilde{G}_{\bar{p}_i}$ has at least one (possibly multiple) correspondence in $G_{\bar{q}_j}$, i.e., 
\begin{equation*}
	\tilde{G}_{\bar{p}_i} = \{\bm{g}^k_{\bar{p}_i}\in \tilde{G}_{\bar{p}_i} |\min_{\bm{g}^l_{\bar{q}_j}\in G_{\bar{q}_j}}\|\hat{\bm{T}}\left(\bm{g}^k_{\bar{p}_i}\right)-\bm{g}^l_{\bar{q}_j}\|_2 < r_p\}.
\end{equation*}
% Each element $\bm{W}^{{\bar{p}_i}}_{kl}$ of the ground-truth matching probability matrix $\bm{W}^{\bar{p}_i}$ between patches $G_{\bar{p}_i}$ and $G_{\bar{q}_j}$ is calculated directly by
% \begin{equation}\label{eq:bn}
% 	\bm{W}^{{\bar{p}_i}}_{kl} = 
% 	\begin{cases}
% 		1, & \|\hat{\bm{T}}\left(G_{\bm{\mu}_{\bar{p}_i}}(k)\right)-G_{\bar{q}_j}(l)\|_2 < r_p \\
% 		0, & \text{otherwise}
% 	\end{cases}.
% \end{equation}
For each anchor $\bm{g}^k_{\bar{p}_i}\in \tilde{G}_{\bar{p}_i}$,  we denote the set of its positive points in $G_{\bar{q}_j}$ as $\mathcal{N}_p^{\bm{g}^k_{\bar{p}_i}}$. All points of $\mathcal{\bm{Q}}$ outside a (larger) radios $r_n$ form the set of its negative patches as $\mathcal{N}_n^{\bm{g}^k_{\bar{p}_i}}$. The fine-level matching loss $\mathcal{L}_{F}^{\mathcal{\bm{P}}}$ on $\mathcal{\bm{P}}$ is calculated as:
\begin{equation}\label{eq:pce}
\begin{aligned}
    \mathcal{L}_{F}^{\mathcal{\bm{P}}} &=\frac{1}{|\bm{\tilde{\mathcal{P}}}|}\sum_{\tilde{\bm{p}}_i\in\bm{\bar{\mathcal{P}}}}\frac{1}{|\tilde{G}_{\bar{p}_i}|} \sum_{\bm{g}^s_{\bar{p}_i}\in \tilde{G}_{\bar{p}_i}}\log\left[1+\xi_s\right], \\
    \xi_s &= \sum_{\bm{g}^k_{\bar{q}_j}\in\mathcal{N}_p^{\bm{g}^s_{\bar{p}_i}}}e^{r^k_s\beta_p^{sk}(d_s^k-\Delta p)}\cdot\sum_{\bm{g}^l_{\bar{q}_j}\in\mathcal{N}_n^{\bm{g}^s_{\bar{p}_i}}}e^{\beta_n^{sl}(\Delta n - d_s^l)},
\end{aligned}
\end{equation}
where $d_s^k=\mathcal{D}_f(\bm{f}_{\bm{g}^s_{\bar{p}_i}},\bm{f}_{\bm{g}^s_{\bar{q}_j}})$ is the distance in the feature space. The weights $\beta_p^{sk}=\omega d_s^k$ and $\beta_n^{sl}=\omega (2.0-d_s^l)$ are determined individually for each positive and negative example with a learned scale factor $\omega\geq 1$. $\Delta p = 0.1$ and $\Delta n = 1.4$. The same goes for the loss $\mathcal{L}_{F}^{\bm{\mathcal{Q}}}$ on $\bm{\mathcal{Q}}$. The
overall superpoint matching loss writes as
\begin{equation}
  \mathcal{L}_{F} = \frac{1}{2}(\mathcal{L}_{F}^{\mathcal{\bm{P}}} + \mathcal{L}_{F}^{\mathcal{\bm{Q}}}).
\end{equation}

\noindent\textbf{Fine-Level Overlap Loss.} The overlap score loss is 
$\mathcal{L}_{FO} = -\frac{1}{2}\left(\frac{1}{|\bm{\mathcal{\bar{P}}}|}\sum_{\bar{p}_i}\mathcal{L}_{\bar{p}_i} + \frac{1}{|\bm{\mathcal{\bar{Q}}}|}\sum_{\bar{q}_j}\mathcal{L}_{\bar{q}_j}\right)$ with
\begin{equation*}\label{eq:lop}
	\mathcal{L}_{\bar{p}_i} = \frac{1}{|\tilde{G}_{\bar{p}_i}|}\sum_{\bm{g}^k_{\bar{p}_i}}\left(\bar{\bm{\mu}}_{\bm{g}^k_{\bar{p}_i}}\log \bm{\mu}_{\bm{g}^k_{\bar{p}_i}}+\left(1-\bar{\bm{\mu}}_{\bm{g}^k_{\bar{p}_i}}\right)\log\left(1-\bm{\mu}_{\bm{g}^k_{\bar{p}_i}}\right)\right).
\end{equation*}
The ground-truth label $\bar{\bm{\mu}}_{\bm{g}^k_{\bar{p}_i}}$ of the point $\bm{g}^k_{\bar{p}_i} \in \tilde{G}_{\bar{p}_i}$ is defined as
\begin{equation}\label{eq:gtop}
	\bar{\bm{\mu}}_{\bm{g}^k_{\bar{p}_i}} = 
	\begin{cases}
		1, & \left(\min_{q_j\in \bm{\mathcal{Q}}}\|\hat{\bm{T}}(\bm{g}^k_{\bar{p}_i}) -\bm{q}_j\|\right) < r_o \\
		0, & \text{otherwise}
	\end{cases},
\end{equation}
where $\mathcal{L}_{\bar{q}_j}$ is calculated in the same way. 

\subsubsection{Implementation Details} 
The proposed method is implemented in PyTorch and can be trained on a single Quadro GV100 GPU (32G)  and two Intel(R) Xeon(R) Gold 6226 CPUs. The hyperparameters are set as follows: $\xi_1=1.0$, $\tau=5.0, \lambda=0.1$, $\epsilon=0.001$, $N_I = 100$, and $N_O=20$. We train our model 120 epochs with a batch size of 1 in all experiments using the ADAM optimizer with an initial learning rate of $5e-4$ and an exponentially decaying factor of 0.99. We adopt the same encoder and decoder architectures used in \cite{yu2021cofinet}. For training the network, we sample 128 coarse correspondences, with truncated patch size $K= 64$ on 3DMatch (3DLoMatch). On KITTI, the numbers are 128 and 32, respectively. 

\section{Experiments}\label{sec:exp}
We conduct extensive experiments to evaluate the performance of our method on indoor 3DMatch \cite{zeng20173dmatch} and 3DLoMatch \cite{huang2021predator} benchmarks, outdoor KITTI \cite{geiger2012we} benchmark, and cross-source 3DCSR \cite{huang2021comprehensive} benchmark.
  
\noindent\textbf{Baselines.} COTReg was compared with the recent state-of-the-art methods: FCGF \cite{choy2019fully}, D3Feat \cite{bai2020d3feat}, SpinNet \cite{ao2021spinnet}, Predator \cite{huang2021predator}, YOHO \cite{wang2021you}, CoFiNet \cite{yu2021cofinet}, and GeoTransformer\cite{qin2022geometric}.

% \review{It is not clear that how the approaches can handle challenging cases, such as 3DLoMatch?}
% \gmei{So in section 5.1, we add the experiments on 3DLoMatch.}
\subsection{Evaluation on 3DMatch and 3DLoMatch.}
\noindent\textbf{Datasets.} 3DMatch \cite{zeng20173dmatch} and 3DLoMatch \cite{huang2021predator} are two widely used indoor datasets that contain more than $30\%$ and $10\% \sim 30\%$ partial overlapping scene pairs, respectively. 3DMatch contains 62 scenes, from which we use 46 scenes for training, 8 scenes for validation, and 8 scenes for testing, respectively. The test set contains 1,623 partially overlapped point cloud fragments and their corresponding transformation matrices. We use training data preprocessed by \cite{huang2021predator} and evaluate on both the 3DMatch and 3DLoMatch \cite{huang2021predator} protocols. We first voxel downsample the point clouds with a $2.5cm$ voxel size, then extract different feature descriptors. Following \cite{huang2021predator}, we set $r_o=3.75cm$, $r_p=3.75cm$, and $r_n=10.0cm$, respectively. 

\noindent\textbf{Metrics.} Following Predator\cite{huang2021predator} and CoFiNet \cite{yu2021cofinet}, we evaluate performance with three metrics: (1) \textit{Inlier Ratio} (IR), the fraction of putative correspondences whose residuals are below a certain threshold (i.e., 0.1m) under the ground-truth transformation, (2) \textit{Feature Matching Recall} (FMR), the fraction of point cloud pairs whose inlier ratio is above a certain threshold (i.e., 5\%), and (3) \textit{Registration Recall} (RR), the fraction of point cloud pairs whose transformation error is smaller than a certain threshold (i.e., $RMSE<0.2m$).
\begin{table}[!htbp]
	\centering
	\caption{Results on both 3DMatch and 3DLoMatch datasets under different numbers of samples. }
	\setlength{\tabcolsep}{0.5mm}
	{
		\begin{tabular}{r | c c c c c | c c c c c}
			\toprule
			~ &  \multicolumn{5}{c|}{3DMatch} &  \multicolumn{5}{c}{3DLoMatch} \\
			\# Samples  & 5000 & 2500 & 1000 & 500 & 250 & 5000 & 2500 & 1000 & 500 & 250 \\
			\midrule
			Method & \multicolumn{10}{c}{Inlier Ratio $(\%) \uparrow$} \\
			\hline
			FCGF\cite{choy2019fully}  		& 56.8 & 54.1 & 48.7 & 42.5 & 34.1 & 21.4 & 20.0 & 17.2 & 14.8 & 11.6\\
			D3Feat\cite{bai2020d3feat}  	& 39.0 & 38.8 & 40.4 & 41.5 & 41.8 & 13.2 & 13.1 & 14.0 & 14.6 & 15.0\\
			SpinNet \cite{ao2021spinnet} 	& 47.5 & 44.7 & 39.4 & 33.9 & 27.6 & 20.5 & 19.0 & 16.3 & 13.8 & 11.1 \\
			Predator \cite{huang2021predator}  & 58.0 & 58.4 & 57.1 & 54.1 & 49.3 & 26.7 & 28.1 & 28.3 & 27.5 & 25.8\\
			CoFiNet\cite{yu2021cofinet}		& 49.8 & 51.2 & 51.9 & 52.2 & 52.2 & 24.4 & 25.9 & 26.7 & 26.8 & 26.9\\
			YOHO \cite{wang2021you} 	& 64.4 & 60.7 & 55.7 & 46.4 & 41.2 & 25.9 & 23.3 & 22.6 & 18.2 & 15.0\\
			GeoTransformer\cite{qin2022geometric} & 71.9 & 75.2 & 76.0 & 82.2 & 85.1 
			& 43.5 & 45.3 & 46.2 & 52.9 & 57.7 \\
			COTReg (Ours)		& \bf85.4 & \bf85.7 & \bf86.1 & \bf86.4 & \bf86.9 & \bf54.2 & \bf55.1 & \bf56.3 & \bf57.7 & \bf59.3\\
			\hline
			& \multicolumn{10}{c}{Feature Matching Recall $(\%) \uparrow$} \\
			\hline
			FCGF\cite{choy2019fully} 		& 97.4 & 97.3 & 97.0 & 96.7 & 96.6 & 76.6 & 75.4 & 74.2 & 71.7 & 67.3 \\
			D3Feat \cite{bai2020d3feat}  	& 95.6 & 95.4 & 94.5 & 94.1 & 93.1 & 67.3 & 66.7 & 67.0 & 66.7 & 66.5 \\
			SpinNet \cite{ao2021spinnet} 	& 97.6 & 97.2 & 96.8 & 95.5 & 94.3 & 75.3 & 74.9 & 72.5 & 70.0 & 63.6 \\
			Predator\cite{huang2021predator} & 96.6 & 96.6 & 96.5 & 96.3 & 96.5 & 78.6 & 77.4 & 76.3 & 75.7 & 75.3\\
			CoFiNet\cite{yu2021cofinet}	& 98.1 & 98.3 & 98.1 & 98.2 & 98.3 & 83.1 & 83.5 & 83.3 & 83.1 & 82.6\\
			YOHO\cite{wang2021you} 		& 98.2 & 97.6 & 97.5 & 97.7 & 96.0 & 79.4 & 78.1 & 76.3 & 73.8 & 69.1\\
			GeoTransformer\cite{qin2022geometric} & 97.9 & 97.9 & 97.9 & 97.9 & 97.6 & 88.3 & 88.6 & 88.8 & 88.6 & 88.3 \\
			COTReg (Ours)   & \bf98.5 & \bf98.6 & \bf98.5 & \bf98.6 & \bf98.6 & \bf89.5 & \bf89.7 & \bf89.7 & \bf89.6 & \bf89.4\\
			\hline
			& \multicolumn{10}{c}{Registration Recall $(\%) \uparrow$} \\
			\hline
			FCGF\cite{choy2019fully} 	& 85.1 & 84.7 & 83.3 & 81.6 & 71.4 & 40.1 & 41.7 & 38.2 & 35.4 & 26.8 \\
			D3Feat\cite{bai2020d3feat} 	& 81.6 & 84.5 & 83.4 & 82.4 & 77.9 & 37.2 & 42.7 & 46.9 & 43.8 & 39.1 \\
			SpinNet\cite{ao2021spinnet} & 88.6 & 86.6 & 85.5 & 83.5 & 70.2 & 59.8 & 54.9 & 48.3 & 39.8 & 26.8 \\
			Predator\cite{huang2021predator} 	& 89.0 & 89.9 & 90.6 & 88.5 & 86.6 & 59.8 & 61.2 & 62.4 & 60.8 & 58.1 \\
			CoFiNet\cite{yu2021cofinet}	  & 89.3 & 88.9 & 88.4 & 87.4 & 87.0 & 67.5 & 66.2 & 64.2 & 63.1 & 61.0 \\
			YOHO \cite{wang2021you} 	  & 90.8 & 90.3 & 89.1 & 88.6 & 84.5 & 65.2 & 65.5 & 63.2 & 56.5 & 48.0 \\
			GeoTransformer\cite{qin2022geometric} & 92.0 & 91.8 & 91.8 & 91.4 & 91.2 & 75.0 & 74.8 & 74.2 & 74.1 & 73.5 \\
			COTReg (Ours)	& \bf93.1 & \bf92.8 & \bf92.9 & \bf92.5 & \bf92.4 & \bf80.9 & \bf80.4 & \bf79.7 & \bf76.0 & \bf74.6 \\			
			\bottomrule
		\end{tabular}
	}
	\label{table:3dm}
\end{table}

\subsubsection{Inlier Ratio and Feature Matching Recall} 
As the main contribution of COTReg is that we jointly adopt the pointwise and structural matchings to estimate the more correct correspondences, we first check the Inlier Ratio of COTReg, which is directly related to the quality of extracted correspondences. Following \cite{huang2021predator,yu2021cofinet,qin2022geometric}, we report the results with different numbers of correspondences. As shown in
Table \ref{table:3dm} (Top), on Inlier Ratio, COTReg outperforms all the previous methods on both benchmarks, showing an outstanding accuracy improvement. In particular, our COTReg surpasses GeoTransformer, the second best baseline, consistently by $1.8\%\sim 13.5\%$ on 3DMatch and $1.6\%\sim 10.7\%$ 3DLoMatch when the sample number ranges from 250 to 5000, respectively. Furthermore, the fact that sampling fewer correspondences leads to a higher Inlier Ratio indicates that our learned scores are well-calibrated, i.e., higher confidence scores indicate more reliable correspondences. For Feature Matching Recall, in Table \ref{table:3dm} (Middle), our method obtains the best results. Especially on 3DLoMatch, which is more challenging due to low overlap scenarios, our proposed method achieves improvements of at least 0.9\%, demonstrating its effectiveness in low overlap cases. The main difference between our method and other baselines is the correspondence prediction strategy, which jointly considers point-wise and structural matchings equipped with overlap scores. On the contrary, these baseline methods only consider pointwise matching based on feature similarity.

\subsubsection{Registration Recall} 
\begin{figure}[t]
	\centering 
	\includegraphics[width=3.6in]{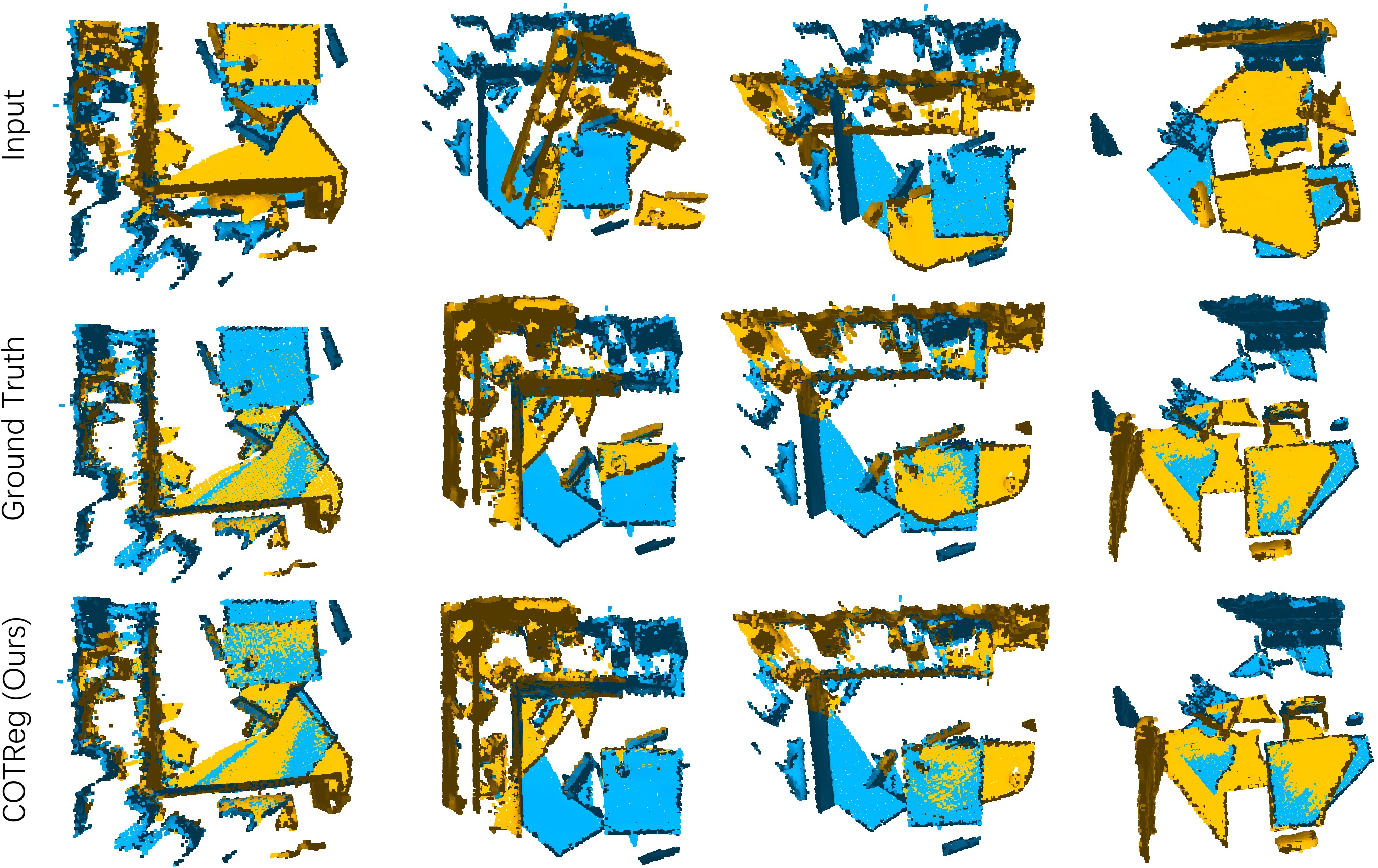}
	\caption{Example qualitative registration results for 3DMatch.}
	\label{fig:3dvs}
\end{figure}

\begin{figure}[t]
	\centering 
	\includegraphics[width=3.6in]{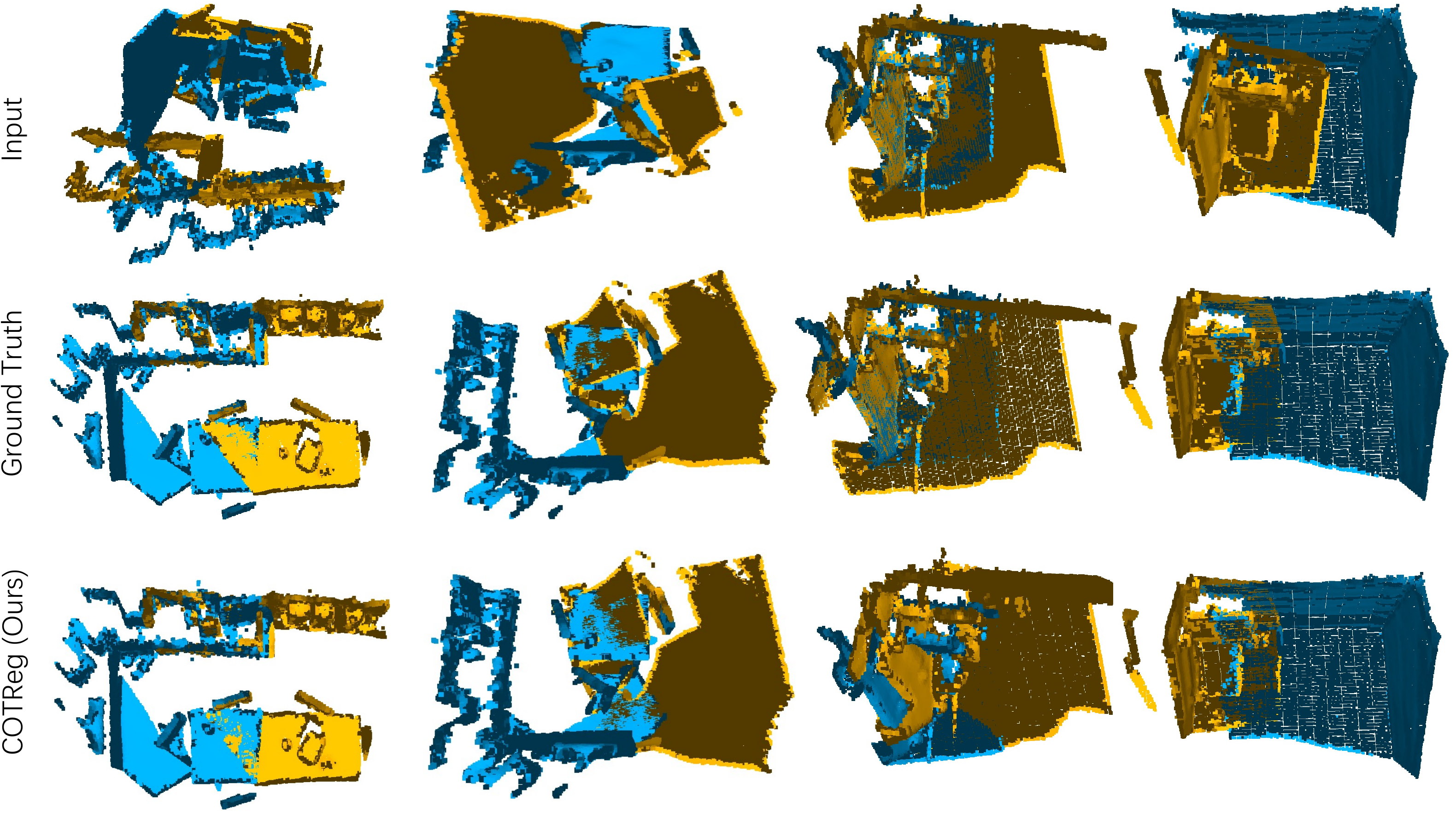}
	\caption{Example qualitative registration results for 3DLoMatch.}
	\label{fig:3dvslo}
\end{figure}

Registration Recall reflects the final performance on point cloud registration. To evaluate the registration performance, we compare the \textbf{RR} obtained by RANSAC in Table \ref{table:3dm} (bottom), and our method outperforms all the other models with various number of sampling
points on both two datasets. Specifically, COTReg achieved $93.1\%$ and $80.9\%$ Registration Recall, exceeding the previous best, GeoTransformer,($ 92.0\% $ RR on 3DMatch) by $ 1.1\% $ and ($ 80.9\% $ RR on 3DLoMatch) by $ 5.9\% $, showing its efficacy in both high- and low-overlap scenarios.
It demonstrates that incorporating both pointwise and structural matchings with overlap scores into the correspondence prediction process can alleviate the ambiguity issue and thus obtains better performance than the counterparts that only consider pointwise matching. Figures \ref{fig:3dvs} and \ref{fig:3dvslo} show visual comparison examples on 3DMatch and 3DLoMatch, respectively. We can easily see that our method can achieve better results in challenging indoor scenes with a low overlap ratio.

% \review{How about solving the weighted least squares for the pose estimation.}
We also compare the registration results weighted SVD over correspondences in Table \ref{tb:3dmsvd}. Some baselines either fail to achieve reasonable results or suffer from severe performance degradation. In contrast, COTReg with weighted SVD achieves the registration recall of 87.2\% on 3DMatch and 60.7\% on 3DLoMatch. Without outlier filtering by RANSAC, high inlier ratio is necessary for successful registration. However, a high inlier ratio does not necessarily lead to a high registration recall, since the correspondences could cluster together as noted in \cite{huang2021predator}. 

% \review{It would be great to include the time comparisons.}
We further count the average inference time of COTReg and compare it with that of the baselines. Notably, all methods consist of two stages that first extract dense features or the correspondences and then recovers the transformation using RANSAC or SVD. We report inference times of the two stages, respectively. Although in the correspondence prediction stage, our method is slightly slower than some baselines, it performs well by extracting reliable correspondences.
%%%%%%%%%%%%%%%%%%%%%%%%%%%%%%%%%%%%%%%%%%%%%%
%%%%%%%%%%%%%%%%%%%%%%%%%%%%%%%%%%%
%%%%%%%%%%%%%%%%%%%%%%%%%
\begin{table}[!htbp]
	\centering
	\caption{Results on both 3DMatch and 3DLoMatch datasets under different numbers of samples. }
	\setlength{\tabcolsep}{0.5mm}
	{
		\begin{tabular}{r | c | c | c c| c c c}
			\toprule
			~ & ~ & ~ & \multicolumn{2}{c|}{RR} &  \multicolumn{3}{c}{Time (s)}\\
			Method & Estimator & Samples & 3DM  & 3DLM & Model & Pose & Total \\
			\midrule
			FCGF\cite{choy2019fully} & RANSAC-50k & 5000 
			& 85.1 & 40.1 & 0.052 & 3.326 & 3.378 \\
			D3Feat\cite{bai2020d3feat}  & RANSAC-50k & 5000 
			& 81.6 & 37.2 & 0.024 & 3.088 & 3.112 \\
			SpinNet \cite{ao2021spinnet}  & RANSAC-50k & 5000 
			& 88.6 & 59.8 & 60.248 & 0.388 & 60.636 \\
			Predator \cite{huang2021predator}  & RANSAC-50k & 5000 
			& 89.0 & 59.8 & 0.032 & 5.120 & 5.152 \\
			CoFiNet\cite{yu2021cofinet}	 & RANSAC-50k & 5000 
			& 89.3 & 67.5 & 0.115 & 1.807 & 1.922 \\
			GeoTrans\cite{qin2022geometric}  & RANSAC-50k & 5000 
			& 92.0 & 75.0 & \bf0.075 & \bf1.558 & \bf1.633 \\
			COTReg (Ours) & RANSAC-50k & 5000 & 
			\bf 93.1 & \bf 80.9 & 0.652 & 1.463 & 2.115 \\
			\hline
			FCGF\cite{choy2019fully} & weighted SVD & 250 
			& 42.1 & 3.9 & 0.052 & 0.008 & 0.056 \\
			D3Feat\cite{bai2020d3feat}  & weighted SVD & 250 
			& 37.4 & 2.8 & 0.024 & 0.008 & 0.032 \\
			SpinNet \cite{ao2021spinnet}  & weighted SVD & 250 
			& 34.0 & 2.5 & 60.248 & 0.006 & 60.254 \\
			Predator \cite{huang2021predator}  & weighted SVD & 250 
			& 50.0 & 6.4 & 0.032 & 0.009 & 0.041 \\
			CoFiNet \cite{yu2021cofinet} & weighted SVD & 250 
			& 64.6 & 21.6 & 0.115 & 0.003 & 0.118 \\
			GeoTrans\cite{qin2022geometric}  & weighted SVD & 250 
			& 86.5 & 59.9 & \bf0.075 & \bf0.003 & \bf0.078 \\
			COTReg (Ours) & weighted SVD & 250
			& \bf87.2 & \bf60.7 & 0.652 & 0.002 & 0.654	\\	
			\bottomrule
		\end{tabular}
	}
	\label{tb:3dmsvd}
\end{table}

\subsection{Evaluation on KITTI}
\subsubsection{Datasets} 
KITTI contains 11 sequences of LiDAR scanned outdoor driving scenarios. For fair comparisons, we follow the same data splitting as \cite{choy2019fully,choy2020deep} and use sequences 0-5 for training, 6-7 for validation, and 8-10 for testing, respectively. Following \cite{choy2020deep}, we further refine the ground truth poses provided using ICP and only use point cloud pairs that are at most $ 10m $ away from each other for evaluation. Following \cite{huang2021predator}, we use a $30cm$ voxel size for downsampling point clouds and set thresholds $r_o$ to $45cm$, $r_p$ to $21cm$, and $r_n$ to $75cm$, respectively.

\subsubsection{Metrics} 
Following Predator~\cite{huang2021predator} and CoFiNet~\cite{yu2021cofinet}, we use three metrics, \textit{Registration Recall} ($RR$), \textit{Rotation Error} ($RRE$), and \textit{Translation Error} ($RTE$), to evaluate the performance of the proposed registration algorithm. $RR$ is the percentage of successful alignment whose rotation error and translation error are below set thresholds (i.e., RRE $<5^\circ$ and RTE $ < 2m $). RRE and RTE are defined as $RE=\arccos\frac{\textbf{Tr}\left(\bm{R}^\top\bm{R}^\star\right)-1}{2}, TE=\|\bm{t}-\bm{t}^\star\|_2 $, respectively. $\bm{R}^\star$ and $\bm{t}^\star$ denote the ground-truth rotation matrix and the translation vector, respectively.

\subsubsection{Registration results} 
We compare to the state-of-the-art RANSAC-based methods: FCGF \cite{choy2019fully}, D3Feat \cite{bai2020d3feat}, SpinNet \cite{ao2021spinnet}, Predator \cite{huang2021predator}, CoFiNet \cite{yu2021cofinet}, and GeoTransformer \cite{qin2022geometric}. As shown in Table \ref{table:kitti}, our model still achieves the best performance in terms of registration recall as well as the lowest average $ RTE $ and $RRE$. The results verify the effectiveness of considering both pointwise and structural matchings to produce correspondences.
\begin{table}[!htbp]
	\centering
	\caption{Results on KITTI dataset. Best performance is highlighted in bold.}
	\begin{tabular}{r | c | c c c}
		\toprule
		Method & Estimator & RTE (cm) $\downarrow$ & RRE $(^\circ)\downarrow$ & RR(\%) $\uparrow$ \\
		\midrule
		FCGF \cite{choy2019fully} & RANSAC 	& 9.5 & 0.30 & 96.6 \\
		D3Feat \cite{bai2020d3feat} & RANSAC 	 & 7.2 & 0.30 & \bf99.8 \\
		SpinNet \cite{ao2021spinnet} & RANSAC & 9.9 & 0.47 & 99.1 \\
		Predator \cite{huang2021predator} & RANSAC  & 6.8 & 0.27 & \bf99.8 \\	
		CoFiNet \cite{yu2021cofinet} & RANSAC	   & 8.5 & 0.41 & \bf99.8 \\
		GeoTrans \cite{qin2022geometric} & RANSAC & 7.4 & 0.27 & \bf99.8 \\
		COTReg (Ours) & RANSAC					& \bf4.9 & \bf0.22 & \bf99.8 \\
		\bottomrule
	\end{tabular}
	\label{table:kitti}
\end{table}
% \begin{figure}[t]
% 	\centering 
% 	\includegraphics[width=3.4in]{figures/kdvs}
% 	\caption{Qualitative registration results on KITTI {\color{blue}REPLACED}.}
% 	\label{fig:kdvs}
% \end{figure}

\subsection{Generalization on Cross-source Dataset}
The generalization ability of learning-based registration algorithms is highly required when the point cloud is acquired from different sensors. To validate the generalizability of our model, we experiment on our own Cross Source Dataset (3DCSR) \cite{huang2021comprehensive}. 3DCSR is a challenging dataset for registration due to a mixture of noise, outliers, density difference, partial overlap, and scale variation.

\subsubsection{3DCSR}
This dataset contains two folders: Kinect Lidar and Kinect SFM. Kinect lidar contains 19 scenes from both the Kinect and Lidar sensors, where each scene is cropped into different parts. Kinect SFM consists of 2 scenes from both Kinect and RGB sensors. The RGB images have already been constructed into a point cloud by using the software VSFM. We use the model trained on 3DMatch since the cross-source dataset is captured in an indoor environment. $RR$ is the percentage of successful alignment whose rotation error and translation error are below set thresholds (i.e., RRE $<15^\circ$ and RTE $ < 6m $).
\begin{table}[!hbt]
	\centering
	\caption{Registration results on Cross Source Datasets. Best performance is highlighted in bold.}
	\begin{tabular}{r | c | c c c}
		\toprule
		Method & Estimator & RRE $(^\circ)\downarrow$ & RTE (cm) $\downarrow$ & RR(\%) $\uparrow$ \\
		\midrule
		FCGF \cite{choy2019fully} & RANSAC	 & 7.47 & \bf0.21 & 49.6  \\
		D3Feat \cite{bai2020d3feat} & RANSAC 	 & 6.41 & 0.26 & 52.0 \\
		SpinNet \cite{ao2021spinnet} & RANSAC & 6.56 & 0.24 & 53.5\\
		Predator \cite{huang2021predator} & RANSAC  & 6.26 & 0.27 & 54.6 \\	
		CoFiNet \cite{yu2021cofinet} & RANSAC	   & 5.76 & 0.26 & 57.3 \\
		GeoTrans \cite{qin2022geometric} & RANSAC & 5.60 & 0.24 & 60.2 \\
		COTReg (Ours) & RANSAC					& \bf5.49 & \bf0.21 & \bf63.4 \\
		\bottomrule
	\end{tabular}
	\label{table:3DCSR}
\end{table}

\subsubsection{Registration Results} 
We use FCGF \cite{choy2019fully}, D3Feat \cite{bai2020d3feat}, SpinNet \cite{ao2021spinnet}, Predator \cite{huang2021predator}, CoFiNet \cite{yu2021cofinet}, and GeoTransformer \cite{qin2022geometric}, as the baselines. Table \ref{table:3DCSR} shows that our method obtains the highest accuracies in generalizing the registration ability to real-world cross-source dataset. Specifically, it outperforms the second-best, GeoTransformer, by more than 3.2\% in terms of registration recall (63.4\% vs 60.2\%). However, the recall is not high enough, showing that registration challenges on 3DCSR remain.
\begin{figure}[t]
	\centering 
	\includegraphics[width=3.4in]{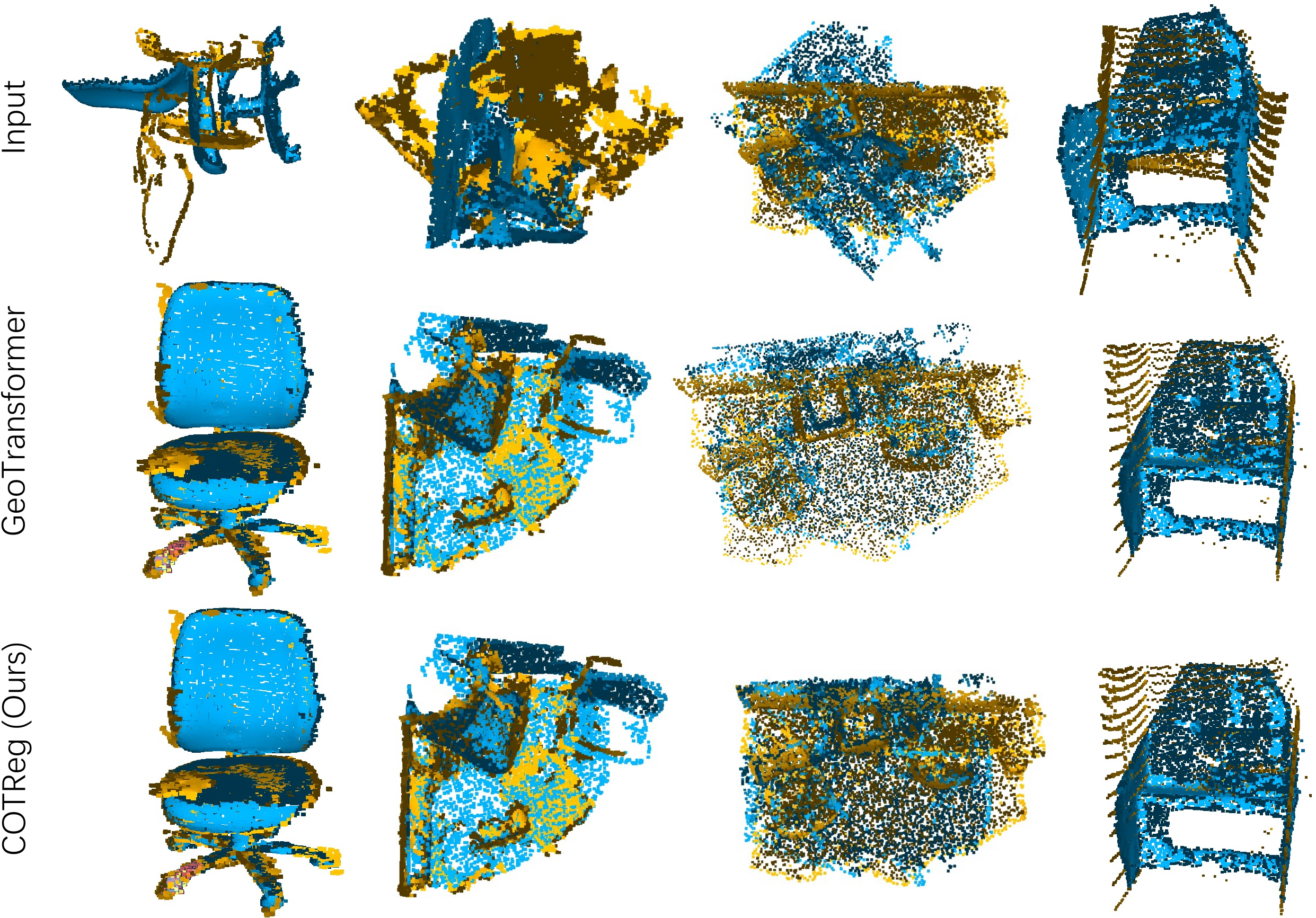}
	\caption{Qualitative registration results on cross source dataset.}
	\label{fig:csdvs}
\end{figure}

\subsection{Ablation Study}
To fully understand COTReg, we conduct an ablation study on 3DMatch and 3DLoMatch to investigate the contribution of each part. First, we replace the overlap scores with a uniform distribution, i.e.,  treating the points in overlap and non-overlap regions equally, to evaluate the effectiveness of overlap scores. As shown in Table \ref{tb:abuot}, on 3DMatch, the learned overlap scores improve the performance by nearly 2.0\%  (92.9\% vs. 90.9\%) RR, 0.7\%  (98.5\% vs. 97.8\%) FMR, and 7.8\%  (86.1\% vs. 68.3\%) IR, respectively. Structure matching can boost RR by 1.1\%  (92.9\% vs. 91.8\%), FMR by 0.5\% (98.5\% vs. 98.0\%) and IR by 10.2\% (86.1\% vs. 75.9\%), respectively. It also indicates that COTReg benefits from the overlap scores and structure matching.  Table \ref{tb:abuot} also shows that the positional encoding can improve the performance in terms of RR, FMR and IR. On 3DLoMatch, the same results can be concluded. 
\begin{table}[!hbt]
	\centering
	\caption{Ablation study of individual modules, tested with \#Samples=1000. \textbf{PM} and \textbf{SM} indicate point matching and structure matching, respectively.  \textbf{OS} indicates point endow with overlap scores. \textbf{PE} represents positional embedding.}
	\small
	\setlength{\tabcolsep}{1.8mm}{
		\begin{tabular}{c c c c| c c c | c c c}
			\toprule
			& & & & \multicolumn{3}{c}{3DMatch} & \multicolumn{3}{c}{3DLoMatch} \\
			\hline
			PE & OS & PM & SM & RR & FMR & IR & RR & FMR & IR \\
			\hline
			$\checkmark$ & $\checkmark$ & $\checkmark$ & $\checkmark$ & \bf92.9 & \bf98.5 & \bf86.1 & \bf79.7 & \bf89.7 & \bf55.1 \\
			$\checkmark$ & $\checkmark$ & $\checkmark$ &  & 91.8 & 98.0 & 75.9 & 74.6 & 88.9 & 46.4 \\ 
			& $\checkmark$ & $\checkmark$ & $\checkmark$		   & 90.9 & 97.8 & 68.3 & 67.2 & 85.6 & 35.4 \\ 
			& $\checkmark$ & $\checkmark$ &  & 90.2 & 97.6 & 63.4 & 66.1 & 84.7 & 33.5 \\ 
			& 			   & $\checkmark$ & $\checkmark$ & 89.6 & 97.6 & 62.9 & 65.0 & 84.3 & 32.1 \\ 
			& & $\checkmark$ &			   & 88.9 & 97.5 & 59.8 & 64.8 & 84.0 & 30.8 \\
			\bottomrule
		\end{tabular}
	}
	\label{tb:abuot}
\end{table}

% Next, we ablate the the hyperparameters of the COTReg. We compare four loss functions for supervising the super- point matching: (a) cross-entropy loss [27], (b) weighted cross-entropy loss [41], (c) circle loss [30], and (d) overlap- aware circle loss. For the first two models, an optimal transport layer is used to compute the matching matrix as in [41]. Circle loss works much better than the two vari- ants of cross-entropy loss, verifying the effectiveness of su- pervising superpoint matching in a metric learning fashion. Our overlap-aware circle loss beats the vanilla circle loss by a large margin on all the metrics.
% \begin{figure}[t]
% 	\centering 
% 	\includegraphics[width=3.4in]{figures/csdvs}
% 	\caption{Qualitative registration results on cross source dataset.}
% 	\label{fig:csdvs1}
% \end{figure}

% \begin{figure}[t]
% 	\centering 
% 	\includegraphics[width=3.4in]{figures/csdvs}
% 	\caption{Qualitative registration results on cross source dataset.}
% 	\label{fig:csdvs2}
% \end{figure}

% \begin{figure}[t]
% 	\centering 
% 	\includegraphics[width=3.4in]{figures/csdvs}
% 	\caption{Qualitative registration results on cross source dataset.}
% 	\label{fig:csdvs3}
% \end{figure}

\section{Conclusion}\label{seq:con}
We propose a method to improve the accuracy of putative correspondences of point cloud registration. Specifically, we design a coupled optimal transport-based approach to generate correspondences by jointly considering overlap scores, pointwise and structural matchings. Comprehensive experiments show that our approach achieves a new state-of-the-art. We believe our techniques have the potential in the applications that require more accurate of 3D point cloud registration.

\ifCLASSOPTIONcaptionsoff
  \newpage
\fi



%\bibliographystyle{IEEEtran}
\bibliographystyle{IEEEtran} 
\bibliography{IEEEfull}


%\end{IEEEbiography}
%% insert where needed to balance the two columns on the last page with
%%\newpage


%\vfill


\newpage
$~~~~~~~~~~~~~~$
\newpage
\appendices
\section{Proof of the Theorem \ref{thm:point}}\label{ap:thm1}
\begin{proof} 
	We assume the correct matches in correspondence set are free of noise. 
	We denote $h_1(\Gamma)=\left<\Gamma,\bm{C}^{pq}\right>$.  
	
	As the minimum values of $h_1(\Gamma)$ is non-negative, if we can prove that $h_1(\Gamma^\star)=0$, then $\Gamma^\star$ is a optimal solution of (\ref{eq:point}). As $\bm{R}^\star, \bm{t}^\star, \Gamma^\star$ are the optimal solutions of problem (\ref{eq:reg0}), we have  
	\begin{equation*}
		\begin{aligned}
			& \sum_{i=1}^N\sum_{j=1}^{M}\Gamma_{ij}^\star\|\bm{R}^\star\bm{p}_i + \bm{t}^\star-\bm{q}_j\|_2^2=0 \\ 
			\Rightarrow
			& \begin{cases}
				\|\bm{R}^\star\bm{p}_i + \bm{t}^\star-\bm{q}_j\|_2^2=0, & \Gamma_{ij}^\star=1 \\
				\Gamma_{ij}^\star \mathcal{D}_f\left(\bm{f}_{\bm{p}_i},\bm{f}_{\bm{q}_j}\right)=0, & \Gamma_{ij}^\star=0
			\end{cases},
		\end{aligned}
	\end{equation*}
	i.e., when $\Gamma_{ij}^\star=1$, $\bm{p}_i\rightarrow\bm{q}_j$ is an aligned correspondence since $\|\bm{R}^\star\bm{p}_i + \bm{t}^\star-\bm{q}_j\|_2^2=0$. As the features are invariant to rigid transformation, then $\bm{f}_{\bm{p}_i} =\bm{f}_{\bm{q}_j}\Rightarrow \Gamma_{ij}^\star \mathcal{D}_f\left(\bm{f}_{\bm{p}_i},\bm{f}_{\bm{q}_j}\right)=0$. Thus $h_1(\Gamma^\star)=0$. 
\end{proof}

\section{Proof of the Theorem \ref{thm:line}}\label{ap:thm2}
\begin{proof} 
	Denote
	$h_2(\Gamma)=\sum_{ijkl}\Gamma_{ij}\Gamma_{kl}\left(\bm{C}^p_{ik}-\bm{C}^q_{jl}\right)^2$. The minimum values of $h_2(\Gamma)$ is non-negative, so if we can prove that $h_2(\Gamma^\star)=0$, then $\Gamma^\star$ is a optimal solution of  (\ref{eq:structure}).
	If $\Gamma_{ij}^\star=1$ and $\Gamma_{k, l}=1$, then $\|\bm{R}^\star\bm{p}_i + \bm{t}^\star-\bm{q}_j\|_2^2=0$ and $\|\bm{R}^\star\bm{p}_k + \bm{t}^\star-\bm{q}_l\|_2^2=0\Rightarrow \|\bm{p}_i -\bm{p}_k\|_2=\|\bm{R}^\star\bm{p}_i + \bm{t}^\star-\bm{R}^\star\bm{p}_k + \bm{t}^\star\|=\|\bm{q}_j -\bm{q}_l\|_2$. Meantime, we can get that $\bm{f}_{\bm{p}_i}=\bm{f}_{\bm{p}_k}$ and $\bm{f}_{\bm{q}_j} =\bm{f}_{\bm{q}_l}$. Then we have 
	\begin{equation*}
		\begin{aligned}
			\left(\bm{C}^p_{ik}-\bm{C}^q_{jl}\right)^2 
			=&[(1-\lambda) \mathcal{D}_f\left(\bm{f}_{\bm{p}_i},\bm{f}_{\bm{p}_k}\right)+\lambda\mathcal{D}_e\left(\bm{p}_i,\bm{p}_k\right)\\
			-&(1-\lambda) \mathcal{D}_f\left(\bm{f}_{\bm{q}_j},\bm{f}_{\bm{q}_l}\right) - \lambda \mathcal{D}_e\left(\bm{q}_j,\bm{q}_l\right)]^2 \\
			=& 0 \Rightarrow \Gamma_{ij}\Gamma_{kl}\left(\bm{C}^p_{ik}-\bm{C}^q_{jl}\right)^2=0. 
		\end{aligned}
	\end{equation*}
	If $\Gamma_{ij}^\star=0$ or $\Gamma_{k, l}=0$, thus $h_2(\Gamma^\star)=0$.
\end{proof}

\section{Proof of the Theorem \ref{thm:dual}}\label{ap:thm3}
\begin{proof}
We denote 
\begin{equation}
	\begin{aligned}
		f(\Gamma^{(k)})&=\xi_1\bm{C}^{pq} +\xi_2\bm{H}\left(\bm{C}^p, \bm{C}^q, \Gamma^{(k)}\right) - \epsilon\log \Gamma^{(k)}, \\
		\mathcal{H}(\Gamma) &= \sum_{i=1}^N\sum_{j=1}^M\Gamma_{ij}(\log\Gamma_{ij}-1).
	\end{aligned}
\end{equation}
And then we have 
\begin{equation*}
	\begin{aligned}
		\mathcal{KL}\left(\Gamma|\Gamma^{(k)}\right)	& = \sum_{i=1}^N\sum_{j=1}^{M} \left(\Gamma_{ij}\log\left(\frac{\Gamma_{ij}}{\Gamma^{(k)}_{ij}}\right)-\Gamma_{ij} + \Gamma^{(k)}_{ij}\right)\\
		& = \mathcal{H}(\Gamma) - \left<\log\Gamma^{(k)},  \Gamma\right> + \bm{1}^\top_N\Gamma^{(k)}\bm{1}_M.
	\end{aligned}	
\end{equation*}
After algebraic simplification, Eq. \eqref{eq:update} can be rewritten as 
\begin{equation}\label{eq:iter11}
	\begin{aligned}
		\Gamma^{(k+1)} & = \mathop{\arg\min}_{\Gamma\geq 0} \left<f\left(\Gamma^{(k)}\right),  \Gamma\right> + \epsilon\mathcal{H}(\Gamma)  \\
		& + \tau\left(\mathcal{KL}\left(\Gamma\bm{1}_M |\bm{\mu}_p\right) + \mathcal{KL}\left(\Gamma^\top\bm{1}_N |\bm{\mu}_q\right)\right)\\
		& + \epsilon\bm{1}^\top_N\Gamma^{(k)}\bm{1}_M,
	\end{aligned}
\end{equation}
with initialization $\Gamma^{(0)}=\bm{\mu}_p{\bm{\mu}_q}^\top$. We solve it iteratively with the help of the Sinkhorn-Knopp algorithm \cite{cuturi2013sinkhorn,chizat2018scaling}. 
For $\forall \epsilon > 0$, we can check that the problem (\ref{eq:iter11}) is strongly convex and lower semi-continuous. Meanwhile, $\bm{\mu}_p$ and $\bm{\mu}_q$ are given non-negative vectors, strong duality and the existence of a minimizer for (\ref{eq:iter11}) is thus given by the Fenchel-Legendre dual form, which states that
	\begin{equation*}\label{eq:dual1}
		\max_{\bm{u}\in\mathbb{R}^N,\bm{v}\in\mathbb{R}^M} -F^\star\left(-\bm{u}\right) 
		-G^\star\left(-\bm{v}\right) - \epsilon \sum_{ij}\exp\left(\frac{u_i+v_j-\bm{C}_{ij}}{\epsilon}\right),
	\end{equation*}
	where $\bm{C}_{ij} = [f\left(\Gamma^{(n)}\right)]_{ij} $, and the function $F^\star\left(\cdot\right)$ and $G^\star\left(\cdot\right)$ take the following
	forms:
	\begin{equation}\label{eq:dual2}
		\begin{aligned}
			F^\star\left(\bm{u}\right) &= \sup_{\bm{z}\in\mathbb{R}^N}\bm{z}^\top\bm{u} - \tau \mathcal{KL}\left(\bm{z}|\bm{\mu}_p\right) \\
			& =\tau\left\langle\exp\left(\frac{\bm{u}}{\tau}\right),\bm{\mu}_p\right\rangle - {\bm{\mu}_p}^\top\bm{1}_N,\\
			G^\star\left(\bm{v}\right) &= \sup_{\bm{z}\in\mathbb{R}^M}\bm{z}^\top\bm{v} - \tau \mathcal{KL}\left(\bm{z}|\bm{\mu}_q\right) \\
			&=\tau\left\langle\exp\left(\frac{\bm{v}}{\tau}\right),\bm{\mu}_q\right\rangle - {\bm{\mu}_q}^\top\bm{1}_M,\\
		\end{aligned}
	\end{equation}
	Thus, it is proved by denoting $h(\bm{u}, \bm{v})=F^\star\left(-\bm{u}\right) 
	+G^\star\left(-\bm{v}\right) + \epsilon \sum_{ij}\exp\left(\frac{\bm{u}_i+\bm{v}_j-\bm{C}_{ij}}{\epsilon}\right)$. 
\end{proof}

\end{document}